\documentclass[times, review, 10pt]{elsarticle}

\usepackage[utf8]{inputenc}
\usepackage{url}
\usepackage{subcaption}
\usepackage{pifont}
\usepackage{pgffor}
\usepackage{nicefrac}
\usepackage{microtype}
\usepackage{adjustbox}
\usepackage{float}
\usepackage[section]{placeins}
\usepackage{orcidlink}
\usepackage{booktabs}
\usepackage{multirow}
\usepackage[table]{xcolor}
\usepackage{amssymb}
\usepackage{wrapfig}
\usepackage{array}
\usepackage{graphicx}
\usepackage{makecell}

\definecolor{BlueGreen}{HTML}{008080}
\definecolor{OliveGreen}{RGB}{128, 128, 0}

\definecolor{mydarkdarkgreen}{RGB}{93, 150, 74}
\definecolor{mydarkgreen}{RGB}{216, 233, 199}
\definecolor{mylightgreen}{RGB}{245, 249, 241}
\definecolor{RedOrange}{RGB}{255, 165, 0}
\definecolor{lightgray}{gray}{.9}

\journal{Journal of \LaTeX\ Templates}

\begin{document}

\begin{frontmatter}

\title{ROVR-Open-Dataset: A Large-Scale Real-World Depth Dataset for Robust Autonomous Driving Perception}

\author[1]{Xianda Guo, \orcidlink{0000-0003-2822-4690}\fnref{fn1}}

\author[2]{Ruijun Zhang, \orcidlink{0009-0006-0788-0604}\fnref{fn1}}

\author[3]{Yiqun Duan, \orcidlink{0000-0003-1517-994X}\fnref{fn1}}

\author[2]{Ruilin Wang, \orcidlink{0009-0009-3567-5263}}
\author[4]{Matteo Poggi, \orcidlink{0000-0002-3337-2236}}
\author[5]{Keyuan Zhou, \orcidlink{0009-0003-0116-0516}}
\author[6]{Wenzhao Zheng, \orcidlink{0000-0001-7188-3734}}
\author[1]{Wenke Huang, \orcidlink{0000-0003-4819-293X}}
\author[7]{Gangwei Xu, \orcidlink{0009-0007-0435-4534}}
\author[8]{Yanlun Peng, \orcidlink{0009-0009-6972-7690}}

\author[9]{Yuan Si, \orcidlink{0009-0004-1513-0438}}


\author[1]{Qin Zou, \orcidlink{0000-0001-7955-0782}\corref{cor1}}
\ead{qzou@whu.edu.cn} 

\cortext[cor1]{Corresponding author.}
\fntext[fn1]{Equal contribution.}

\affiliation[1]{organization={School of Computer Science, Wuhan University}, country={China}}
\affiliation[2]{organization={Institute of Automation, Chinese Academy of Sciences (CASIA)}, country={China}}
\affiliation[3]{organization={University of Technology Sydney}, country={Australia}}
\affiliation[4]{organization={University of Bologna}, country={Italy}}
\affiliation[5]{organization={Zhejiang University}, country={China}}
\affiliation[6]{organization={University of California, Berkeley}, country={USA}}
\affiliation[7]{organization={Huazhong University of Science and Technology}, country={China}}
\affiliation[8]{organization={Great Wall Motor}, country={China}}
\affiliation[9]{organization={ROVR Labs, Inc.}, country={USA}}

\begin{abstract}
Depth estimation is a fundamental component of spatial perception for autonomous driving and other unmanned systems operating in open urban environments. Existing depth datasets such as KITTI, nuScenes, and DDAD have advanced the field but are limited in diversity and scalability, and benchmark performance on them is approaching saturation. A less discussed constraint is \emph{sensor economics}: the bespoke multi-LiDAR rigs behind these datasets are expensive, power-hungry, and difficult to replicate at fleet scale, which caps the geographic and temporal diversity that any single benchmark can cover.
We present ROVR, a large-scale, diverse, and cost-efficient depth dataset designed to capture the complexity of real-world driving. ROVR comprises 200K high-resolution frames across highway, rural, and urban scenarios, spanning day/night cycles and adverse weather conditions, collected across North America, Europe, and Asia. 
We additionally release the calibration, synchronization, preprocessing, and privacy pipeline so that the platform can be reproduced by third parties.

Extensive ablation studies further characterize performance across scene types, illumination, weather conditions, and ground-truth sparsity levels, and identify three qualitatively distinct failure modes---photometric collapse, and geometric confusion---that current architectures share. Beyond monocular depth estimation itself, ROVR provides realistic geometric supervision for spatial embodied intelligence in open urban environments, where robust depth is a prerequisite for navigation, mapping, and multimodal scene understanding in unmanned systems. Together these results establish ROVR as a demanding benchmark for developing depth estimation models with stronger real-world robustness.
The dataset and code are publicly available at \url{https://xiandaguo.net/ROVR-Open-Dataset}.

\end{abstract}



\end{frontmatter}

\section{Introduction}
\label{sec:intro}

Depth estimation is a cornerstone of 3D scene understanding, providing essential geometric information for a wide range of applications, including autonomous driving~\cite{duan2023diffusiondepth,guo2023openstereo,ldvls2023}, robotics~\cite{guo2025lightstereo, stereoanything,horc2021}, and augmented reality (AR)~\cite{westermeier2024assessing}. Accurate depth maps are essential for downstream tasks such as obstacle detection, motion planning, semantic mapping, and 3D reconstruction, making depth estimation a bottleneck for the entire autonomous driving stack. Over the past decade, the availability of high-quality annotated datasets has been a major driver of progress, enabling increasingly sophisticated learning-based approaches.

Notable examples include large-scale driving datasets such as KITTI~\cite{kittidepth}, nuScenes~\cite{nuscenes}, and DDAD~\cite{Guizilini2020ddad}, which have served as critical benchmarks for monocular and stereo depth estimation. However, despite their impact, each presents limitations in scene diversity, annotation density, sensor calibration, or scalability. KITTI~\cite{kittidepth}, while offering high-precision LiDAR depth, is restricted to a narrow set of urban and suburban scenarios, and its limited scale constrains the training of high-capacity models. nuScenes~\cite{nuscenes} broadens the range of environments and weather conditions but yields sparser and lower-quality ground truth due to its 32-beam LiDAR. DDAD~\cite{Guizilini2020ddad} achieves dense and long-range depth measurements via a sophisticated multi-sensor setup, yet its reliance on costly hardware makes large-scale collection economically prohibitive.

A factor that is often left implicit in dataset papers, but which determines what is actually achievable at fleet scale, is the \emph{cost structure of the sensing platform}. A mechanical spinning 64-beam LiDAR of the kind deployed by KITTI, or the multi-LiDAR rigs behind DDAD and recent research platforms, cost one to two orders of magnitude more per unit than current automotive-grade solid-state LiDARs and require a correspondingly larger enclosure, power budget, and maintenance overhead. As a result, the geographic and temporal coverage of a dataset is ultimately bounded by the number of vehicles that can be instrumented with research-grade hardware---a bound that the community is already bumping against, as modern foundation models outgrow the diversity of existing benchmarks. Solid-state LiDAR, global-shutter HDR cameras, and globally available RTK GNSS corrections have recently matured to the point where a commodity, automotive-grade sensor suite can match the calibration precision of a research rig while being cheap enough to replicate across continents. ROVR exploits exactly this shift.

With the rapid evolution of deep neural networks---particularly transformer-based architectures~\cite{zhao2022monovit, guo2023simple} and multi-modal foundation models---there has been a significant surge in demand for large-scale, diverse, and challenging datasets. As model capacity grows, performance on existing benchmarks is approaching saturation, leaving diminishing room for improvement without fundamentally new data sources. This trend is especially evident in autonomous driving, where domain-specific datasets can no longer keep pace with the generalization requirements of modern models. In the era of foundation models and multi-modal learning, there is an urgent need for a new generation of depth datasets that can scale to support robust spatial understanding.

We present ROVR, a new \emph{large-scale, diverse, and cost-efficient} public dataset for depth estimation in dynamic outdoor environments. This release contains 200K high-resolution video frames spanning a wide variety of scenes---highway, rural, and urban---and conditions---day, night, and rain---while maintaining low capture cost through a lightweight acquisition pipeline. Compared to existing datasets, our ground-truth depth is \emph{sparse} but statistically sufficient for training robust depth estimators. The lightweight pipeline is designed for extensibility, enabling future scaling in both volume and scene diversity with minimal additional cost. We additionally document every component of the pipeline---from the optical and electrical specifications of the sensor suite through calibration, synchronization, privacy blurring, and quality control---so that the platform can be reproduced outside of a single laboratory, which we believe is a prerequisite for any dataset that aims to keep pace with foundation-model scaling.

Beyond scale, ROVR introduces \emph{new challenges} for the community. First, its driving diversity is significantly broader than KITTI or DDAD, requiring models to generalize across heterogeneous sampling distributions. Second, the sparser ground truth imposes a harder learning problem. As a result, state-of-the-art models trained on KITTI suffer severe degradation when evaluated on ROVR (Table~\ref{tab:k_d_n_Rovr}), and even models trained directly on ROVR remain far from saturation (Table~\ref{tab:kitti_rovr}).

In summary, our contributions are:
\begin{itemize}
    \item We introduce ROVR, a large-scale depth dataset of 200K frames collected across three continents, covering diverse scenes, weather, and illumination conditions with a cost-efficient acquisition pipeline. The full dataset, data loaders, calibration and privacy pipelines, and evaluation code are publicly released (Table~\ref{tab:stereo_datasets}, Figure~\ref{fig:cars}).
    \item We document a \emph{reproducible}, commodity-class sensor suite---solid-state LiDAR, global-shutter HDR camera, triple-frequency RTK GNSS, tactical-grade IMU---and the full calibration, synchronization, and privacy pipeline that turns it into a research-grade acquisition platform, including the engineering trade-offs that distinguish it from mechanical multi-LiDAR rigs.
    \item We establish a cross-domain evaluation protocol: models trained on existing benchmarks fail on ROVR, and ROVR-trained models are evaluated back on KITTI, nuScenes, and DDAD to assess transfer in both directions.
\end{itemize}

\begin{table}[t]
\centering
\caption{\textbf{Comparison of available autonomous driving  depth datasets.}}
\label{tab:stereo_datasets}
\fontsize{8pt}{9.5pt}\selectfont
\setlength\tabcolsep{4pt}
\renewcommand\arraystretch{1.1}
\begin{tabular}{l|c|rr|c|c}
\toprule
\rowcolor{gray!20}
Dataset & Scenario & \multicolumn{2}{c|}{Frames} & Resolution & Labels \\
\midrule

KITTIDepth~\cite{kittidepth}   &Driving&86,000&7,000& 1226 $\times$ 370 & \checkmark \\
\rowcolor{gray!10}

\rowcolor{gray!10}
nuScenes~\cite{nuscenes}  &Driving&34,149&5,851&1600 $\times$ 900 & \checkmark \\

DDAD~\cite{Guizilini2020ddad}  & Driving& 17,050&4,150& 1936$\times$1216& \checkmark \\

\rowcolor{gray!10}

\rowcolor{cyan!10}
\textbf{ROVR (Ours)} &Driving& 193,648&10,002& 1920$\times$1280& \checkmark \\
\bottomrule
\end{tabular}
\end{table}

\section{Related Work}
\label{sec:related}

\noindent\textbf{Autonomous Driving Depth Datasets.}
The \textit{KITTI} benchmark~\cite{kittidepth} is one of the earliest and most widely used datasets for driving-scene depth estimation, providing stereo imagery and sparse LiDAR ground truth up to $\sim$80\,m (85K/7K, $1226\times370$). It mainly covers urban and highway scenes under fair weather, limiting diversity in environmental conditions and viewpoints. KITTI has defined the evaluation protocol for the field for more than a decade, but recent state-of-the-art methods have effectively saturated its leaderboard with $\delta_1>0.98$, leaving little headroom to drive further modelling progress.
\textit{nuScenes}~\cite{nuscenes} extends to a 360$^\circ$ multi-camera setup with a 32-beam LiDAR, radar, and GPS/IMU, collected in Boston and Singapore (34K/5.8K, $1600\times900$). It offers broader variety---including night and rain---but yields sparser per-image depth due to lower-resolution LiDAR, and its geographic coverage is limited to two cities.
\textit{DDAD}~\cite{Guizilini2020ddad} addresses range and density constraints via long-range (250\,m) high-density LiDAR and high-resolution cameras, providing near-dense depth for diverse urban scenes across multiple continents (17K/4.1K, $1936\times1216$). DDAD's platform, however, relies on a custom multi-LiDAR rig that is expensive and hard to replicate, which in turn caps the achievable fleet size.
\textit{Waymo Open}~\cite{waymo} provides 600K frames at $1920\times1280$ using a multi-LiDAR and multi-camera rig, but its licensing restricts free redistribution of derived assets, and only a fraction of sequences include per-pixel depth. In short, existing driving depth datasets force a trade-off between density (DDAD, Waymo), diversity (nuScenes), and scale (KITTI). ROVR is positioned to relax all three constraints simultaneously by adopting a lightweight, replicable sensor suite rather than a bespoke research rig.

\noindent\textbf{Other Depth Datasets.}
Beyond autonomous driving, depth datasets span indoor, outdoor, mixed, and synthetic domains.
For indoor scenes, \textit{NYUv2}~\cite{silberman2012nyu} (795/654 frames, $640\times480$), \textit{ScanNet}~\cite{dai2017scannet} (2.5M frames), and \textit{TUM RGB-D}~\cite{sturm2012rgbd} provide RGB-D benchmarks, while \textit{Booster}~\cite{Ramirez2022,Ramirez2023} targets challenging transparent and reflective objects.
For mixed environments, \textit{MegaDepth}~\cite{li2018megadepth} ($\sim$128K frames via multi-view stereo), \textit{DIODE}~\cite{diode_dataset}, and \textit{DA-2K}~\cite{depthanythingv2} offer diverse scenes with dense depth.
Synthetic datasets such as \textit{SceneNet RGB-D}~\cite{mccormac2017scenenet} (5M frames, $320\times240$) enable controlled pre-training and domain adaptation, but the sim-to-real gap remains the bottleneck for deployment in driving.
Additional driving-oriented datasets include \textit{Cityscapes}~\cite{cordts2016cityscapes}, \textit{RobotCar}~\cite{RobotCar}, and \textit{DrivingStereo}~\cite{yang2019drivingstereo} (174K/7.7K frames, $1762\times800$).
Compared to these efforts, ROVR provides 193K/10K high-resolution ($1920\times1280$) driving frames with centimetre-level GNSS pose accuracy, combining the scale of Waymo with broader scene coverage and a lightweight acquisition pipeline that is reproducible outside a single laboratory.

\noindent\textbf{Depth Sensing Hardware and Platform Design.}
The choice of depth sensor directly constrains both the statistics of the resulting dataset and the economics of acquisition. Early autonomous-driving datasets (KITTI) employed mechanical spinning LiDARs---e.g., the Velodyne~HDL-64E---which offer uniform 360$^{\circ}$ coverage but suffer from high cost, significant power draw, and mechanical wear that makes fleet-scale deployment fragile. Second-generation platforms such as nuScenes adopted mid-tier spinning units (Velodyne~VLP-32C), reducing cost at the price of per-frame density. DDAD~\cite{Guizilini2020ddad} and more recent research platforms use multi-LiDAR rigs that stitch several high-channel-count units into near-dense depth, which is effective for benchmarking but economically prohibitive at the scale required to capture long-tail driving conditions.
Solid-state LiDARs based on microelectromechanical (MEMS) or flash-array designs are now commercially viable and have become the dominant modality in consumer-class Level-2+ driver assistance systems. Relative to spinning units, they eliminate moving parts, reduce power consumption, and can be sealed into compact IP-rated enclosures, at the cost of a more anisotropic beam pattern and systematic rather than random sparsity on low-reflectance surfaces. These trade-offs are precisely the ones we exploit in the ROVR platform (Sec.~\ref{sec:sensor-suite}); they turn the acquisition rig from a research artefact into something that can be replicated across three continents at fleet scale.
Complementary to LiDAR, modern automotive cameras have converged on global-shutter CMOS with hardware HDR via multi-exposure stacking, providing an operating envelope that covers both deep shadow and direct sunlight in a single frame. When combined with a multi-frequency RTK GNSS receiver and a tactical-grade IMU, these commodity components form a sensor suite whose per-unit cost is roughly two orders of magnitude below a traditional research autonomy rig, but whose calibration and synchronization discipline can match the precision needed for pixel-level depth supervision.
Orthogonal lines of work explore passive depth cues---stereo~\cite{guo2023openstereo,guo2025lightstereo,stereoanything} and structure-from-motion~\cite{li2018megadepth}---as a way to avoid LiDAR altogether. These approaches are attractive for scalability but struggle in the ill-conditioned regions that define the hard cases of driving (low-texture road surfaces, heavy rain, night), which motivates keeping active sensing in the loop while making the sensor cheaper.

\noindent\textbf{Depth for Spatially Embodied Unmanned Systems.}
For unmanned systems deployed in open urban environments, metric depth is not only an evaluation target but also a shared geometric substrate for mapping, localization, obstacle avoidance, and navigation~\cite{dong2022depth4robotics,gsi2024}. From this perspective, a useful benchmark must stress not only pixel-wise accuracy, but also robustness to weather, illumination, geographic shift, and realistic sensor sparsity. ROVR is designed around exactly these pressures, making it relevant to the broader agenda of spatial embodied intelligence in cities rather than to monocular depth estimation in isolation.

\noindent\textbf{Monocular Depth Estimation.}
The introduction of end-to-end trainable neural networks for monocular metric depth estimation (MMDE), pioneered by~\cite{Eigen2014}, marked a milestone by introducing optimization with the scale-invariant log loss ($\mathrm{SI}_{\log}$). Subsequent advances span convolution-based architectures~\cite{Laina2016, Patil2022p3depth}, diffusion-based models~\cite{duan2023diffusiondepth} and transformer-based models~\cite{Yang2021, Bhat2020adabins, Yuan2022newcrf, piccinelli2023idisc, zhao2022monovit,catnet2024}, achieving impressive in-domain accuracy---e.g., PackNet~\cite{Guizilini2020ddad}, DPT~\cite{Ranftl2021dpt}, and AdaBins~\cite{Bhat2020adabins} report 5--6\% relative error on KITTI, approaching LiDAR precision. However, these models often overfit to their training domain and degrade sharply in novel scenes.
To improve cross-domain generalization, recent works explore large-scale and diverse training data, as in the \emph{Depth-Anything} series~\cite{depthanythingv1,depthanythingv2}, or adopt generalizable MMDE strategies~\cite{bhat2023zoedepth, guizilini2023zerodepth, yin2023metric3d} that incorporate camera awareness~\cite{facil2019camconvs, guizilini2023zerodepth} or normalize outputs via intrinsics~\cite{Lee2019bts, Lopez2020mapillary, yin2023metric3d}. While geometric pretraining~\cite{bhat2023zoedepth} and dataset-specific priors~\cite{yin2023metric3d} can boost zero-shot accuracy, these methods typically assume noiseless pinhole intrinsics and rely on predefined backprojection, limiting applicability. Benchmarks such as the Monocular Depth Estimation Challenge (MDEC)~\cite{spencer2023monocular,spencer2023second,spencer2024third,obukhov2025fourth} now track both in-domain and zero-shot performance, with DepthAnything~v2 and Marigold~\cite{marigold} as strong recent baselines.
A limitation shared by this body of work is that evaluation datasets remain dominated by a handful of benchmarks that do not reflect modern deployment conditions. ROVR complements this line of research by offering a dataset whose sensor statistics, geographic coverage, and sparsity profile better match what a commodity depth estimator actually sees in the wild.

\section{ROVR Dataset}
\label{sec:dataset}

In this section, we describe ROVR, a large-scale open depth dataset tailored for monocular depth estimation in autonomous driving. We first motivate the design philosophy of the acquisition platform (Sec.~\ref{sec:design-philosophy}), then detail the sensor suite and its optical/electrical characteristics, followed by the calibration and synchronization protocol (Sec.~\ref{sec:sensor-calib-sync}), the data processing and privacy pipeline (Sec.~\ref{sec:processing}), the data organization and release plan (Sec.~\ref{sec:organization}), and the resulting dataset statistics (Sec.~\ref{sec:statistics}).


\noindent
\begin{minipage}{\linewidth}
    \centering

    \begin{minipage}[t]{0.45\linewidth}
        \centering
        \includegraphics[
            width=\linewidth,
            keepaspectratio,
            trim=0 0 0 2cm,
            clip
        ]{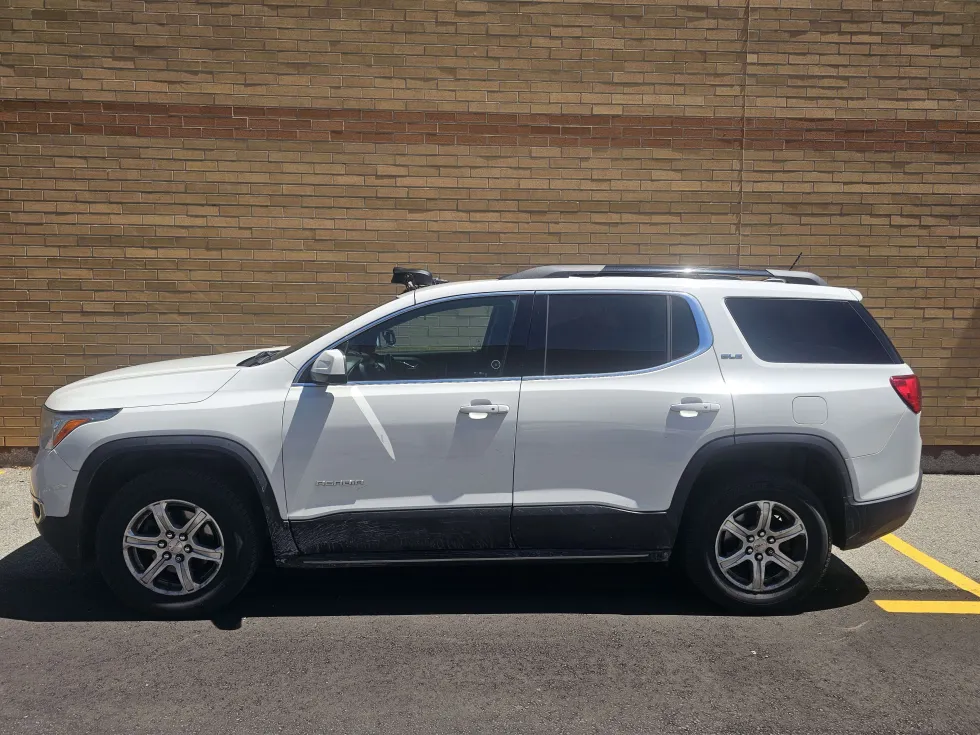}

        \vspace{1mm}
        {\small (a) Real vehicle}
        \label{fig:car2}
    \end{minipage}
    \hfill
    \begin{minipage}[t]{0.45\linewidth}
        \centering
        \includegraphics[width=\linewidth]{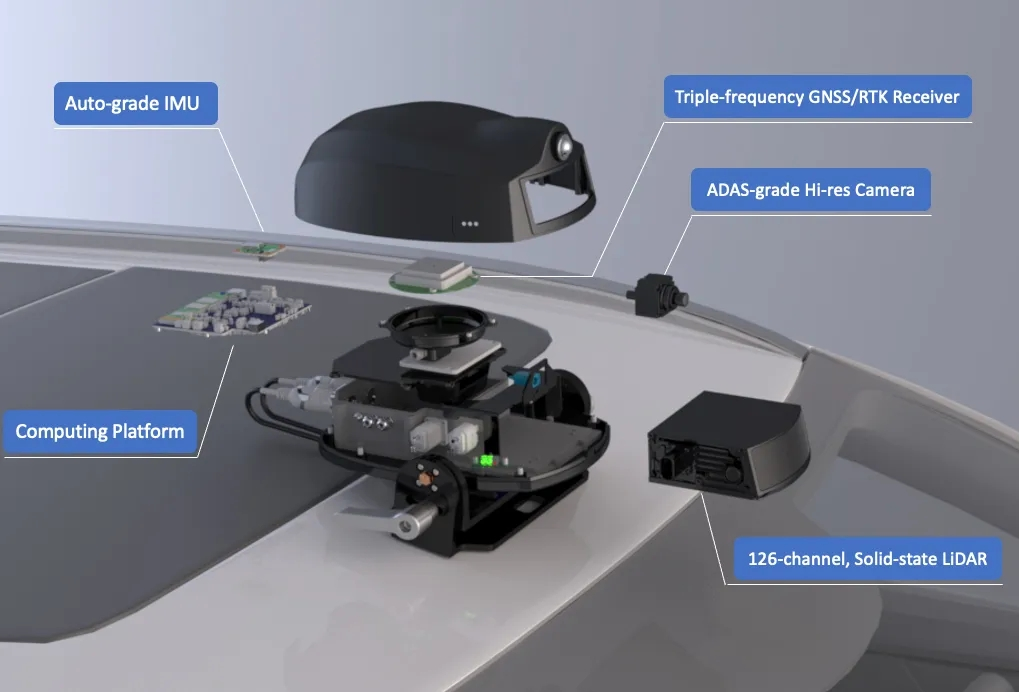}

        \vspace{1mm}
        {\small (b) Sensor suite on vehicle}
        \label{fig:car3}
    \end{minipage}

    \vspace{1mm}
    \captionof{figure}{
        Illustration of the data collection vehicles:
        (a) real-world vehicle used for urban field tests;
        (b) sensor suite on Vehicle~3, including IMU, GNSS/RTK,
        camera, LiDAR, and computing platform.
    }
    \label{fig:cars}
    \vspace{-3mm}
\end{minipage}

\subsection{Design Philosophy: Cost-Efficient, Scalable Sensing}
\label{sec:design-philosophy}
The scalability of a depth dataset is constrained less by algorithms than by \emph{sensor economics}. High-end mechanical LiDAR units of the kind used in KITTI (Velodyne~HDL-64E) or the earliest nuScenes platforms cost an order of magnitude more per unit than current automotive-grade solid-state LiDARs, and their moving parts make long-duration fleet deployment fragile, power-hungry, and difficult to weatherproof. DDAD~\cite{Guizilini2020ddad} addresses range and density through a bespoke multi-LiDAR rig, but the resulting platform is heavy, expensive, and hard to replicate at the scale required to capture the long tail of real-world driving conditions.

ROVR instead adopts a \emph{lightweight, commodity} sensing philosophy. A single automotive-grade solid-state LiDAR, a high-resolution global-shutter RGB camera, a triple-frequency RTK GNSS receiver, and a low-bias-instability IMU are integrated into a compact, fully sealed enclosure that can be mounted on any consumer vehicle. The complete unit draws less than 30\,W under continuous operation and costs roughly two orders of magnitude less than a research-grade autonomy rig. This design choice directly enables the 10{,}000+ hours of driving data aggregated for this release, and it is the reason the acquisition pipeline can be extended across three continents without compromising data quality. The trade-off is that individual LiDAR sweeps are \emph{sparser} than those produced by spinning 64-/128-beam mechanical units. A central empirical question in this paper is therefore whether such cost-efficient but sparser ground truth is sufficient to train competitive depth estimators. Our density ablation (Sec.~\ref{sec:experiments}) answers this affirmatively.

\subsection{Sensor Suite, Calibration, and Synchronization}
\label{sec:sensor-calib-sync}
Each acquisition vehicle carries the same factory-calibrated sensor head in an IP67-rated IP67 enclosure, integrating synchronized LiDAR, RGB, GNSS/INS, IMU, compute, and storage modules. Fig.\ref{fig:cars} shows the external form factor and internal layout.

\paragraph{Sensor Suite.}
\label{sec:sensor-suite}
The primary depth sensor is an automotive-grade 126-beam solid-state LiDAR operating at $905,\mathrm{nm}$, producing $1.2\times 10^{6}$ points per second with a $200,\mathrm{m}$ range at $10\%$ reflectivity and $\pm 2,\mathrm{cm}$ precision at $50,\mathrm{m}$. Its $\pm 12.5^{\circ}\times120^{\circ}$ FOV is aligned with a $1920\times1080$, $30,\mathrm{Hz}$ global-shutter HDR RGB camera equipped with a $3.55,\mathrm{mm}$ lens ($\sim100^{\circ}$ horizontal FOV). Vehicle pose is estimated by a tightly coupled GNSS/INS stack with triple-frequency multi-constellation RTK and a $100,\mathrm{Hz}$ tactical-grade IMU, using GEODNET corrections for centimetre-level localization. On-board synchronization, compression, and privacy blurring run on a $6,\mathrm{TOPS}$ edge AI unit, with data stored on a 1,TB SSD. The actively cooled enclosure operates from $-30^{\circ}\mathrm{C}$ to $+70^{\circ}\mathrm{C}$ across deployment regions.

\paragraph{Calibration.}
Camera intrinsics are estimated per vehicle using a $9\times6$ asymmetric checkerboard observed from at least 60 viewpoints. We fit a Brown--Conrady distortion model and obtain sub-pixel reprojection error ($<0.3$,px RMSE), storing the resulting parameters in \texttt{int.yaml}. LiDAR--camera extrinsics are initialized from CAD and refined by minimizing reprojection error between LiDAR-detected checkerboard corners or pole-like structures and their image detections. The recovered transform $(\mathbf{R}{l\to c},\mathbf{t}{l\to c})$ maps LiDAR points $\mathbf{p}_l$ to the image plane by
\begin{equation}
\mathbf{p}c \sim \mathbf{K} \bigl( \mathbf{R}{l \rightarrow c}, \mathbf{p}l + \mathbf{t}{l \rightarrow c} \bigr),
\label{eq:lidar_to_cam}
\end{equation}
where $\mathbf{K}$ is the camera intrinsic matrix. Extrinsics are stored in \texttt{ext.yaml} and re-verified quarterly or after mechanical incidents. Lever-arm offsets among the IMU, GNSS antenna, LiDAR, and camera are initialized from CAD and refined using slow turning manoeuvres, yielding residuals below $2,\mathrm{cm}$.

\paragraph{Synchronization and Motion Compensation.}
All sensors are hardware-synchronized to a GPS-disciplined clock distributed by the RTK receiver. LiDAR and camera timestamps remain within $2,\mathrm{ms}$ of GNSS time. We operate the LiDAR at $5,\mathrm{Hz}$ and subsample the camera stream to $5,\mathrm{Hz}$, aligning the LiDAR scan midpoint with the camera exposure midpoint to reduce motion-induced projection error. Intra-sweep LiDAR motion is compensated by interpolating the $100,\mathrm{Hz}$ IMU-fused pose to the camera exposure midpoint before projection.

\paragraph{Comparison with Prior Platforms.}
Compared with denser but harder-to-scale multi-LiDAR rigs, ROVR uses a single solid-state LiDAR with globally deployable RTK while retaining DDAD/Waymo-class image resolution. The resulting non-uniform angular density and road-centered sparsity are quantified in Sec.\ref{sec:statistics}.

\subsection{Data Processing and Privacy Pipeline}
\label{sec:processing}
Each raw acquisition is passed through an automated pipeline before being admitted to the release. The pipeline is organized as five stages.

\textbf{(1) Deskew and projection.} For every LiDAR sweep, the motion-compensated points are transformed into the camera frame using Eq.~\ref{eq:lidar_to_cam} and rendered into a single-channel $1920\times 1080$ depth map in PNG (16-bit, millimetre units). Points that fall behind the image plane or outside the camera FOV are discarded. The resulting depth map is by construction pixel-aligned with the RGB frame.

\textbf{(2) Invalid-return masking.} A depth return is marked invalid if (a)~its intensity channel is below the noise floor reported by the LiDAR, (b)~its projected pixel lies inside a detected windshield-reflection or lens-flare mask, or (c)~it corresponds to a LiDAR-direct sunlight condition in which the manufacturer's point-validity flag is de-asserted. These pixels are set to zero in the depth map and are masked out of both training and evaluation.

\textbf{(3) Condition tagging.} Each frame is automatically tagged with scene type (highway / rural / urban) based on GPS mapping and road-class lookup, and with condition (normal / night / rain) based on camera exposure statistics, on the state of the vehicle wiper, and on weather-service ground truth queried from historical APIs. Tags are stored per-frame in the metadata and are the basis for the stratified evaluations of Sec.~\ref{sec:experiments}.

\textbf{(4) Privacy anonymization.} Faces and license plates are detected by a pair of off-the-shelf detectors run on the edge compute unit, and their bounding boxes are Gaussian-blurred with a radius proportional to the detected box size. Detections are reviewed manually for a random audit sample of 1\% of frames; no residual personally identifying information was observed in that audit. The blurring operation is applied in place, so no original unblurred RGB leaves the vehicle.

\textbf{(5) Quality control.} Finally, automated per-sequence checks reject any frame with (i)~LiDAR-camera temporal offset above 2\,ms, (ii)~GNSS horizontal uncertainty above 10\,cm, or (iii)~more than 90\% invalid depth pixels. Sequences with more than 5\% rejected frames are quarantined for manual review. Approximately 2.4\% of raw hours are excluded at this step; the 200K released frames are the post-filter set.

\subsection{Data Organization and Release}
\label{sec:organization}
The processed ROVR dataset is released in ROS~2 bag format. Each sequence folder is named \texttt{\textless UTC\_time\textgreater\allowbreak-\textless device\_ID\textgreater\allowbreak-\textless seq\_ID\textgreater\allowbreak-\textless code\textgreater}, and within each sequence, synchronized multi-modal data are stored in modality-specific subdirectories: $1920\times1080$ RGB images (\texttt{images/}), LiDAR point clouds in PCD format (\texttt{pointclouds/}), LiDAR-projected depth maps in single-channel PNG (\texttt{depth/}), and perception annotations (\texttt{annotation/}). Auxiliary navigation data include interpolated GNSS/INS poses (\texttt{ego\_poses.json}), raw GNSS logs (\texttt{ego\_poses\_raw.json}), and $100\,\mathrm{Hz}$ IMU measurements (\texttt{imu\_data.csv}). All files carry nanosecond timestamps. Calibration parameters accompany each sequence, with camera intrinsics in \texttt{int.yaml} and LiDAR-to-camera extrinsics in \texttt{ext.yaml} (Rodrigues rotation vector and translation, with the coordinate remapping described in Sec.~\ref{sec:sensor-calib-sync}).

The full dataset will be publicly released under the CC~BY-NC-SA~4.0 license at \url{https://xiandaguo.net/ROVR-Open-Dataset/}. The release includes: (1)~$1920\times1080$ RGB images; (2)~LiDAR point clouds in PCD format; (3)~LiDAR-projected depth maps in single-channel PNG; (4)~per-frame GNSS/INS poses; (5)~intrinsic and extrinsic calibration files; and (6)~official train/test splits. We additionally provide PyTorch data loaders, evaluation scripts that reproduce every table in this paper, and an online visualization gallery for browsing dataset samples. The full pipeline described in Sec.~\ref{sec:processing}, including the privacy-blurring detectors and quality-control scripts, is released under the same license, so that third parties extending the dataset can replicate the exact preprocessing.

\begin{table}[t]
    \centering
    \setlength\tabcolsep{1.5pt}
    \renewcommand\arraystretch{1.1}
    \caption{\textbf{Statistical overview of the ROVR dataset.}}
    \fontsize{8pt}{9.5pt}\selectfont
    \begin{tabular}{l|ccc|c||ccc|c}
        \toprule
        \rowcolor{gray!20}
        &\multicolumn{4}{c||}{\textbf{Train}} & \multicolumn{4}{c}{\textbf{Test}} \\
        \rowcolor{gray!20} 
        \multirow{-2}{*}{\textbf{Setting}}& \textbf{Highway} & \textbf{Rural} & \textbf{Urban} &\textbf{Sum} &\textbf{Highway} & \textbf{Rural} & \textbf{Urban} &\textbf{Sum}\\ 
        \midrule
        \textbf{Normal} & 114,644 & 6,411 & 42,937 & 163,992 &5,533 & 600 &2,080&8,213 \\
        \rowcolor{gray!10}
        \textbf{Night} & 20,223 & 0 & 1,650 & 21,873 & 1,040&0&150& 1,190 \\
        \textbf{Rain} & 1,496 & 4,041 & 2,246 & 7,783 & 150 &299&150&599 \\

        \midrule
        \rowcolor{cyan!10}
        \textbf{Sum}& 136,393 & 10,452 & 46,833 & 193,648 & 6,723&899&2,380&10,002 \\
        \bottomrule
    \end{tabular}
    \label{tab:stereo_data}
\end{table}

\noindent
\begin{minipage}{\linewidth}
    \centering
    \setlength{\tabcolsep}{2pt}
    \renewcommand{\arraystretch}{1.0}
    \scriptsize
    \begin{tabular}{cccc}
        \includegraphics[width=0.22\textwidth]{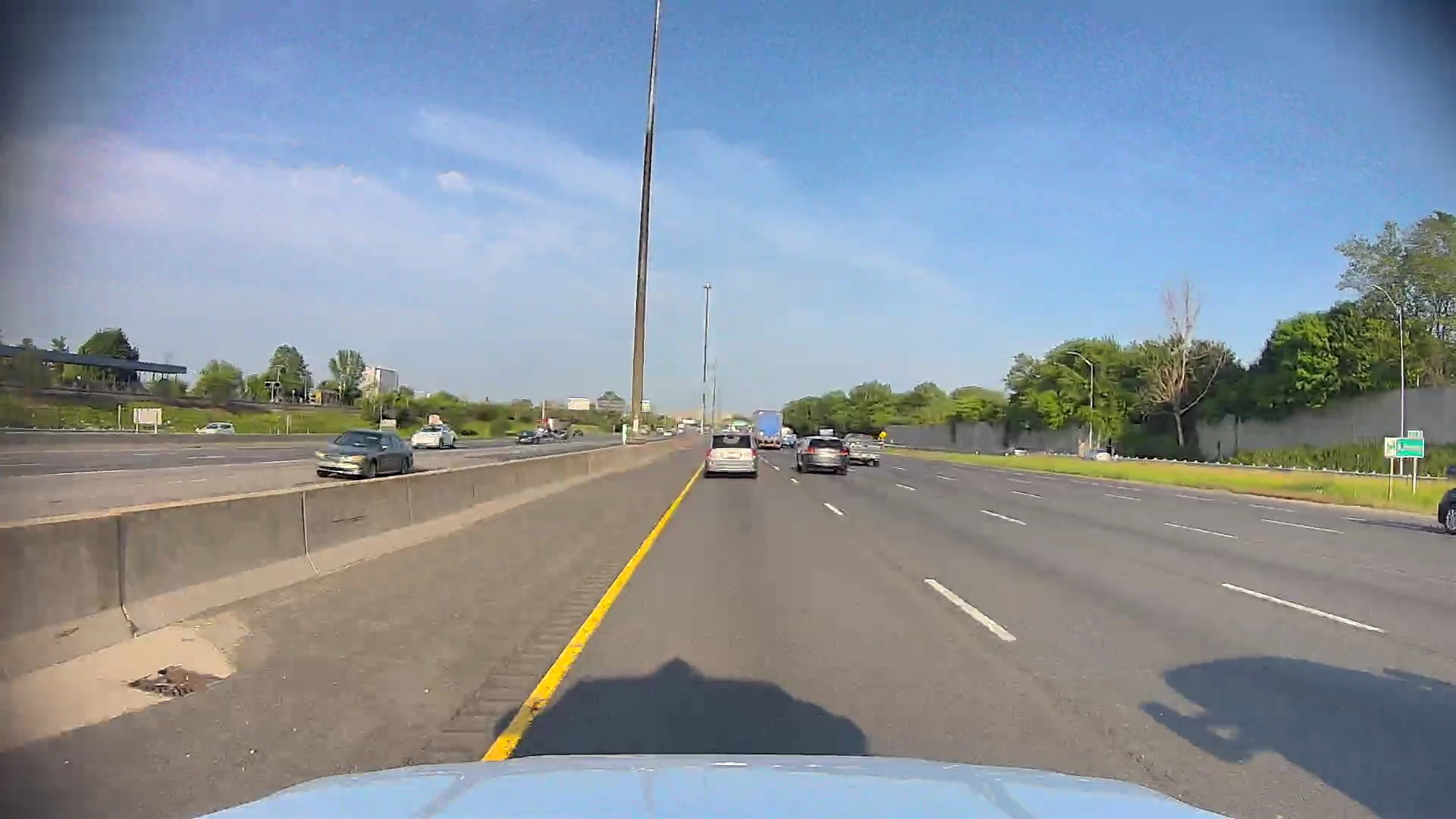} &
        \includegraphics[width=0.22\textwidth]{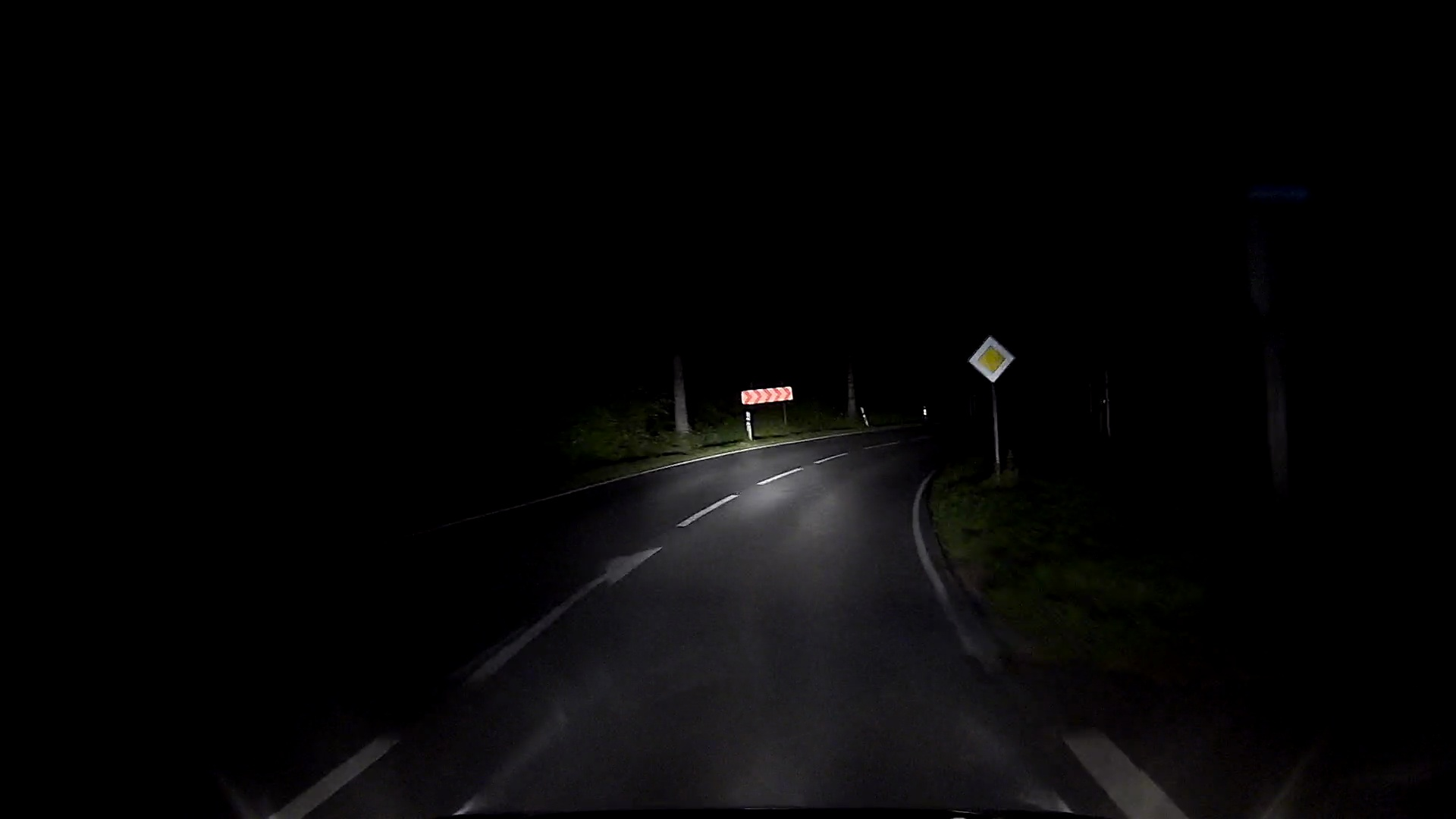} &
        \includegraphics[width=0.22\textwidth]{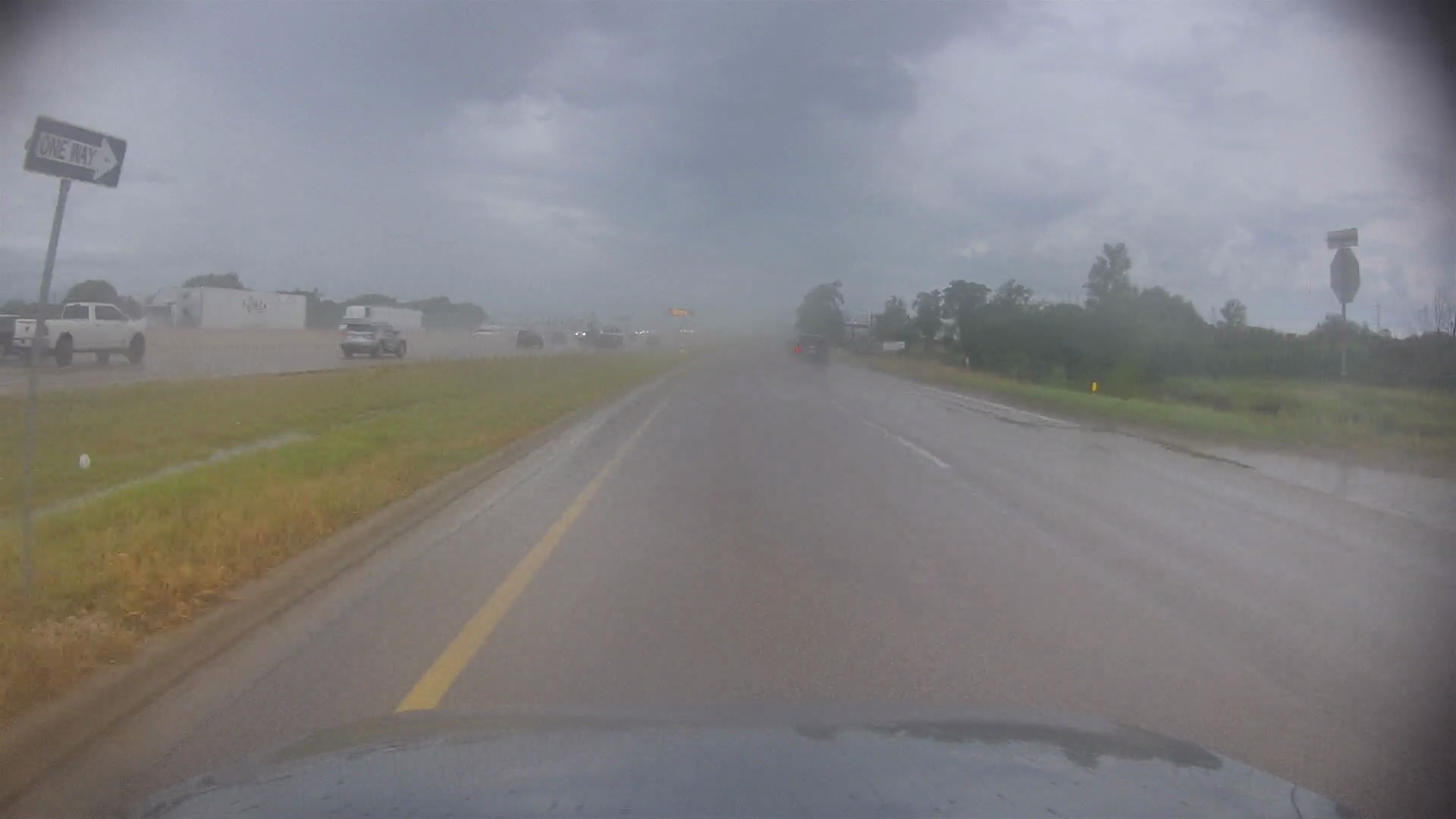} &
        \includegraphics[width=0.22\textwidth]{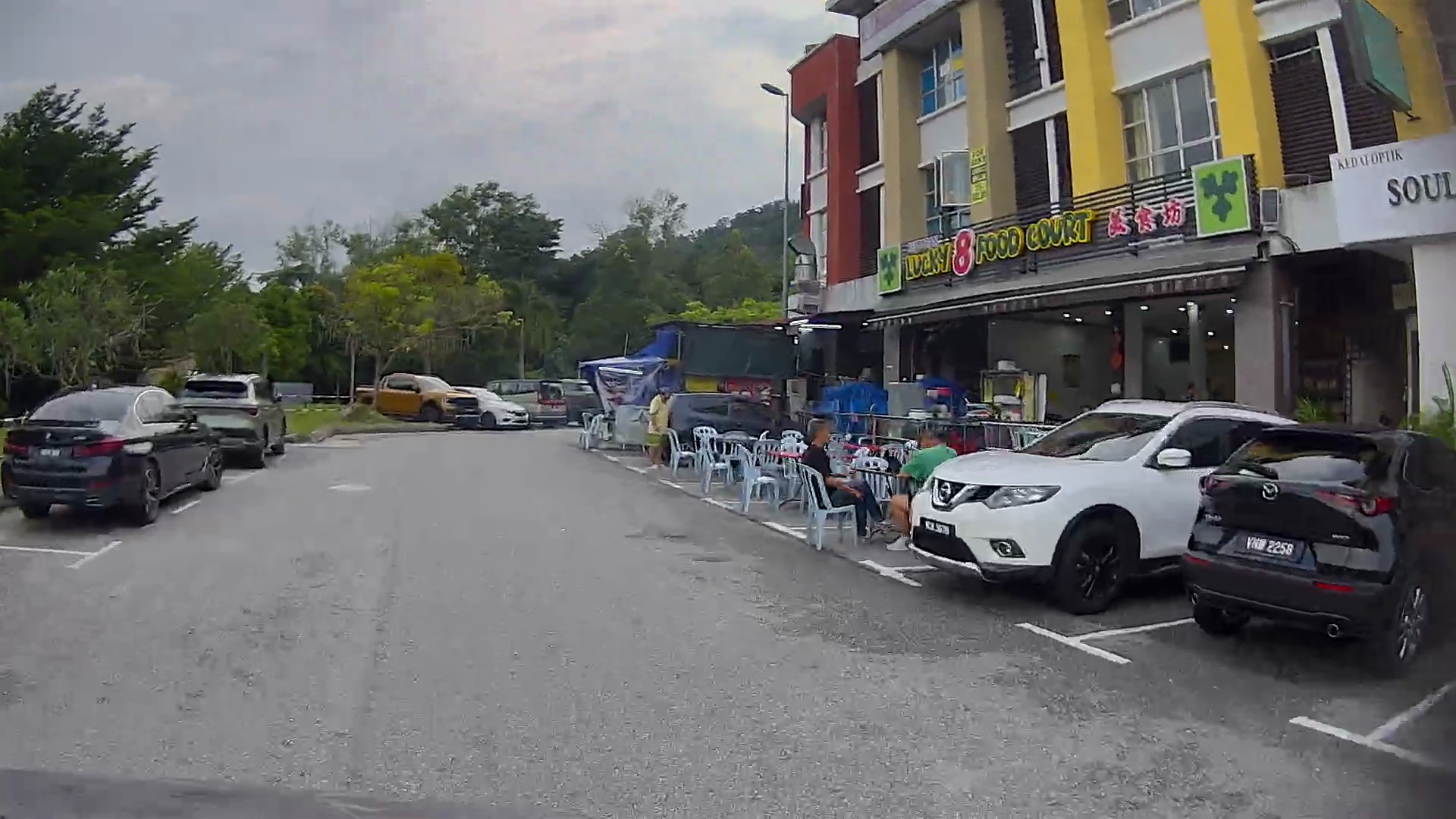} \\

        \includegraphics[width=0.22\textwidth]{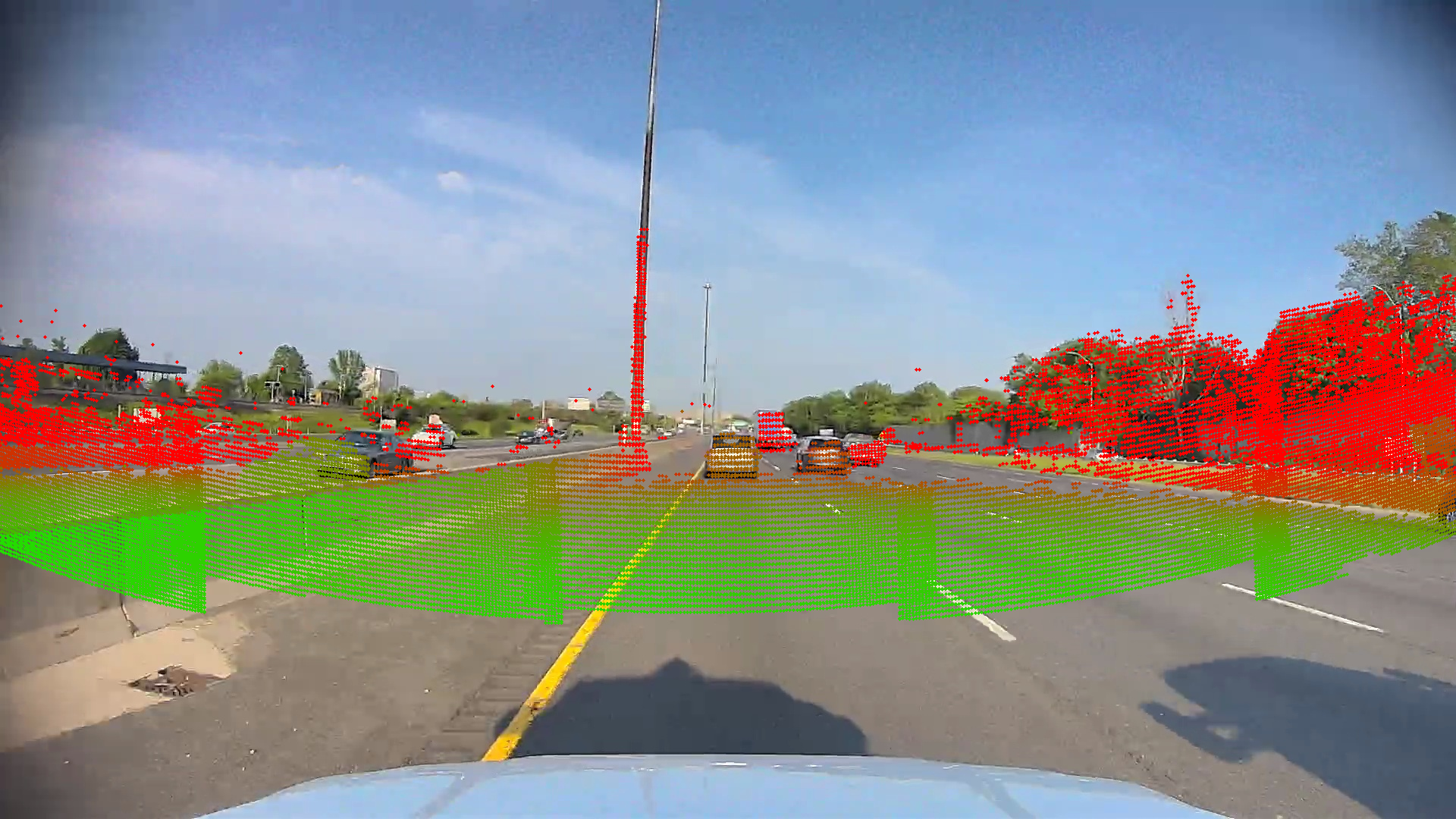} &
        \includegraphics[width=0.22\textwidth]{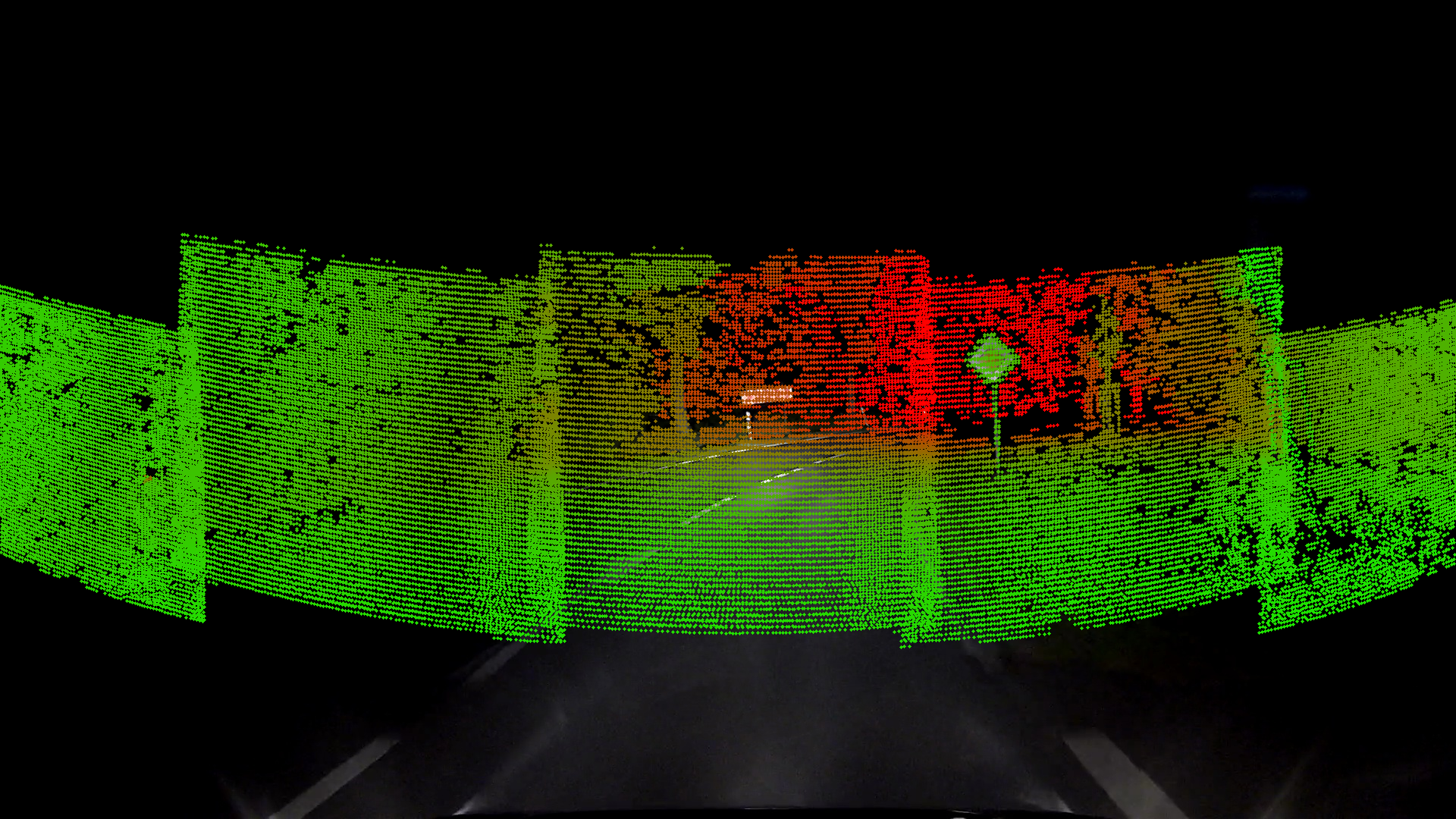} &
        \includegraphics[width=0.22\textwidth]{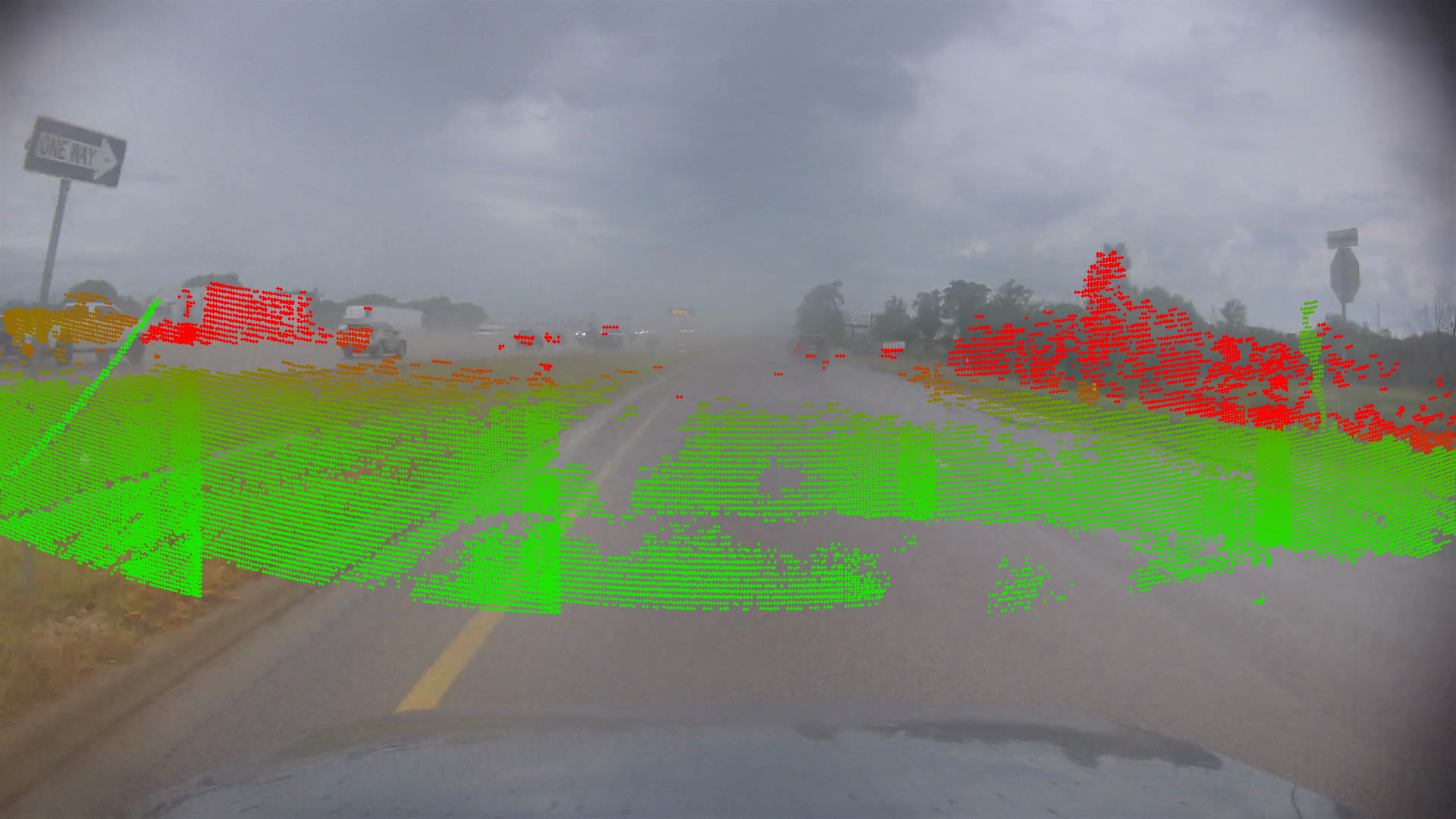} &
        \includegraphics[width=0.22\textwidth]{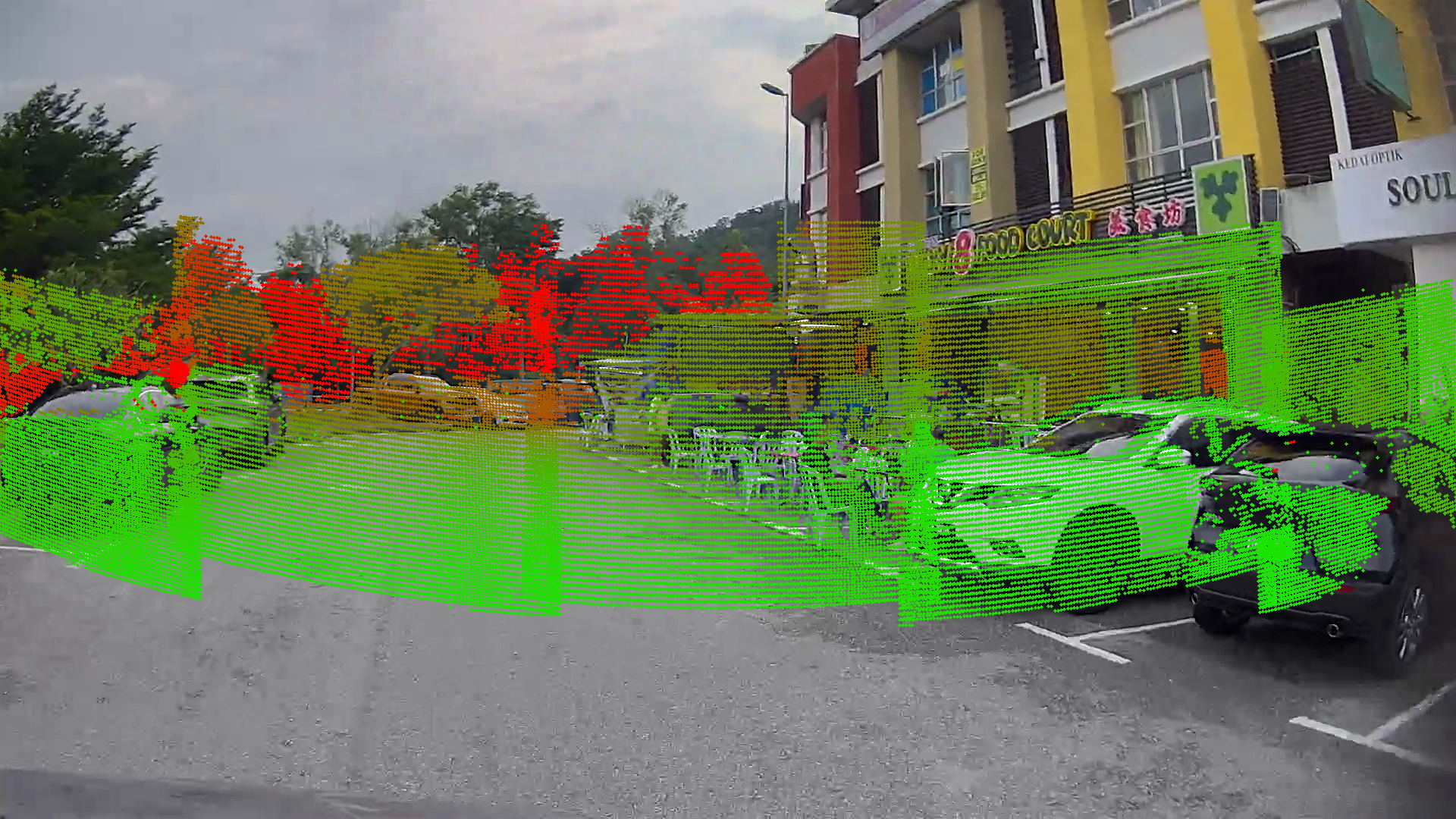} \\

        (a) Highway (Normal) & (b) Highway (Night) & (c) Highway (Rainy) & (d) Urban (Normal) \\[2pt]

        \includegraphics[width=0.22\textwidth]{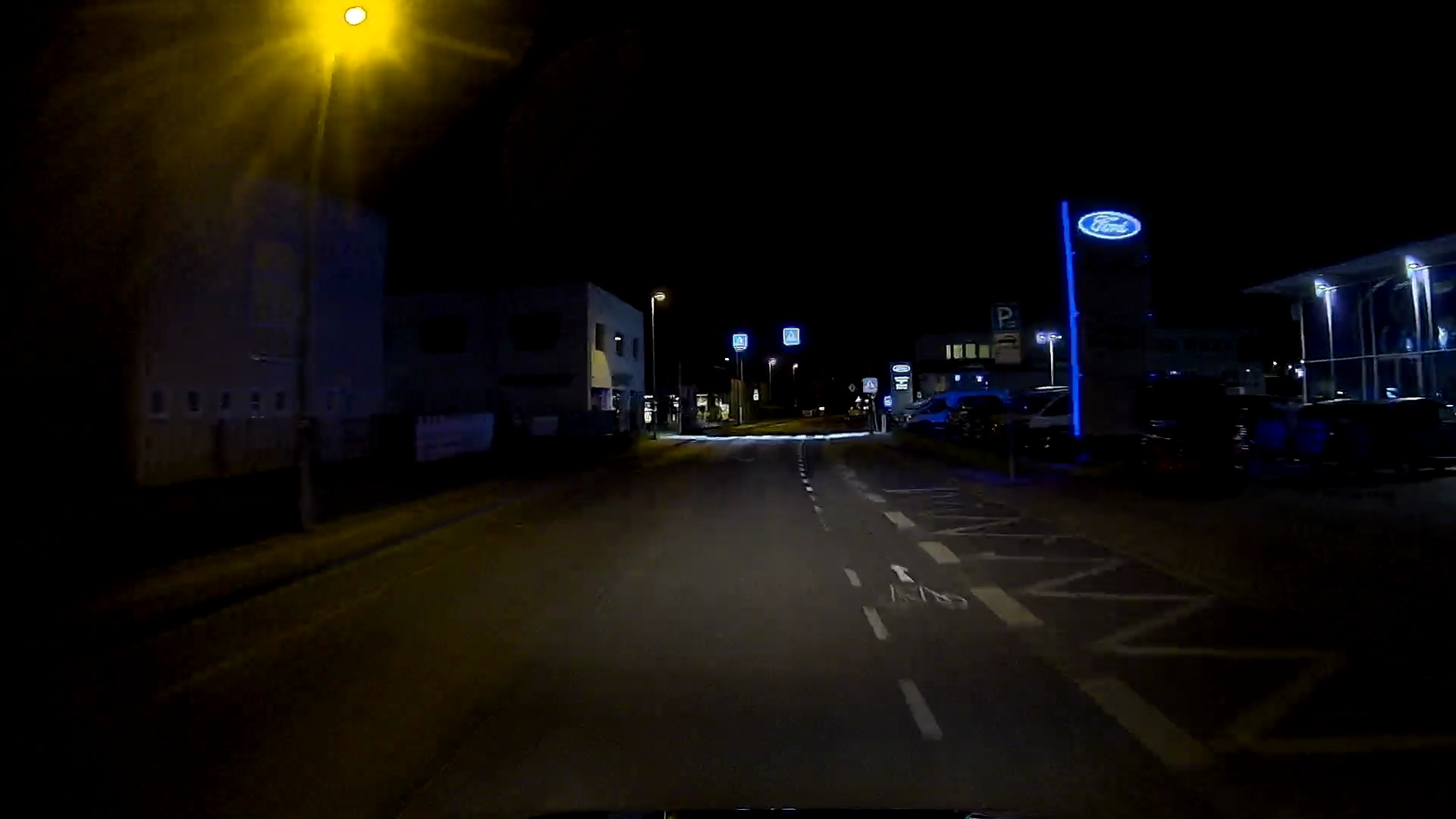} &
        \includegraphics[width=0.22\textwidth]{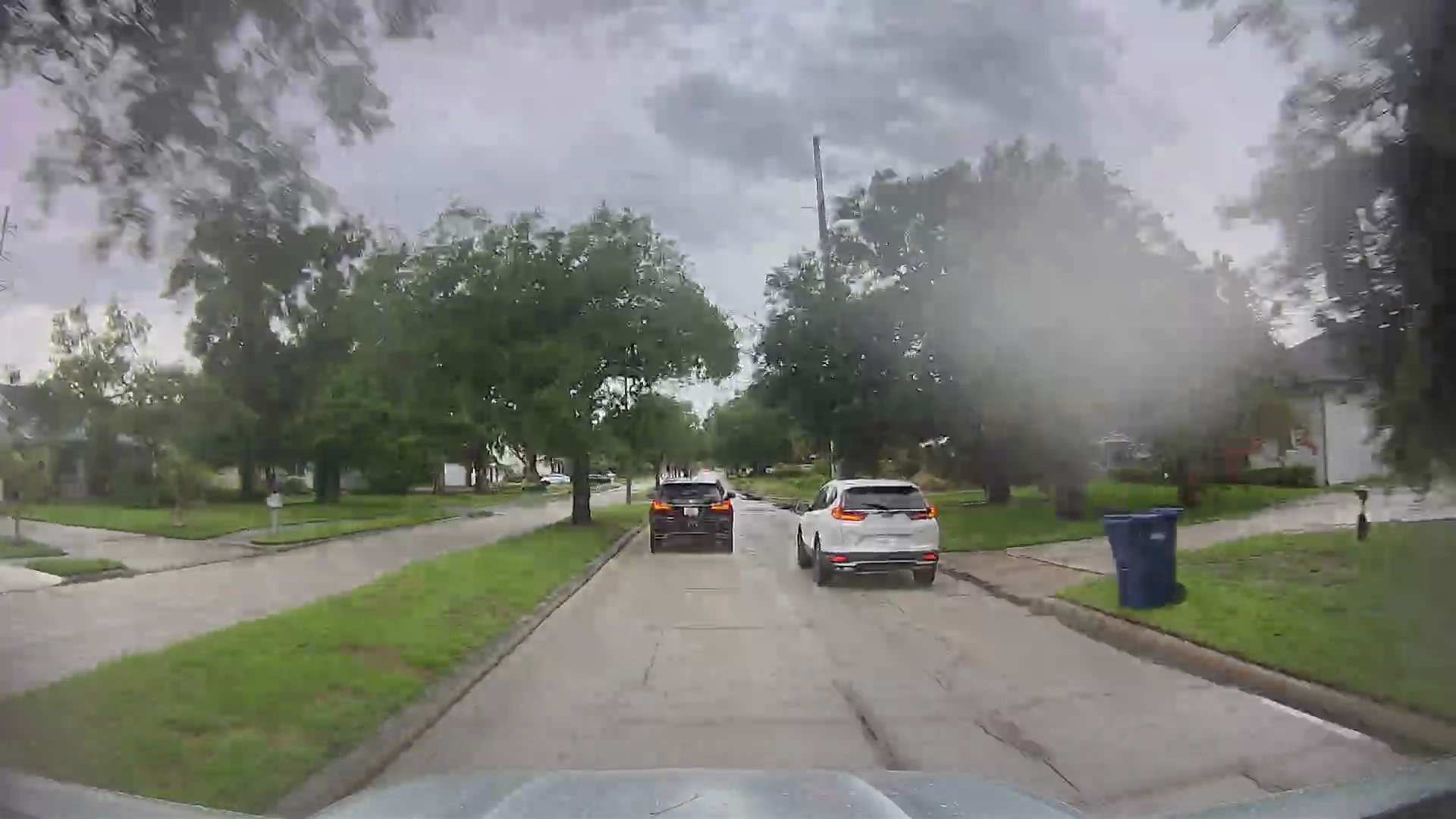} &
        \includegraphics[width=0.22\textwidth]{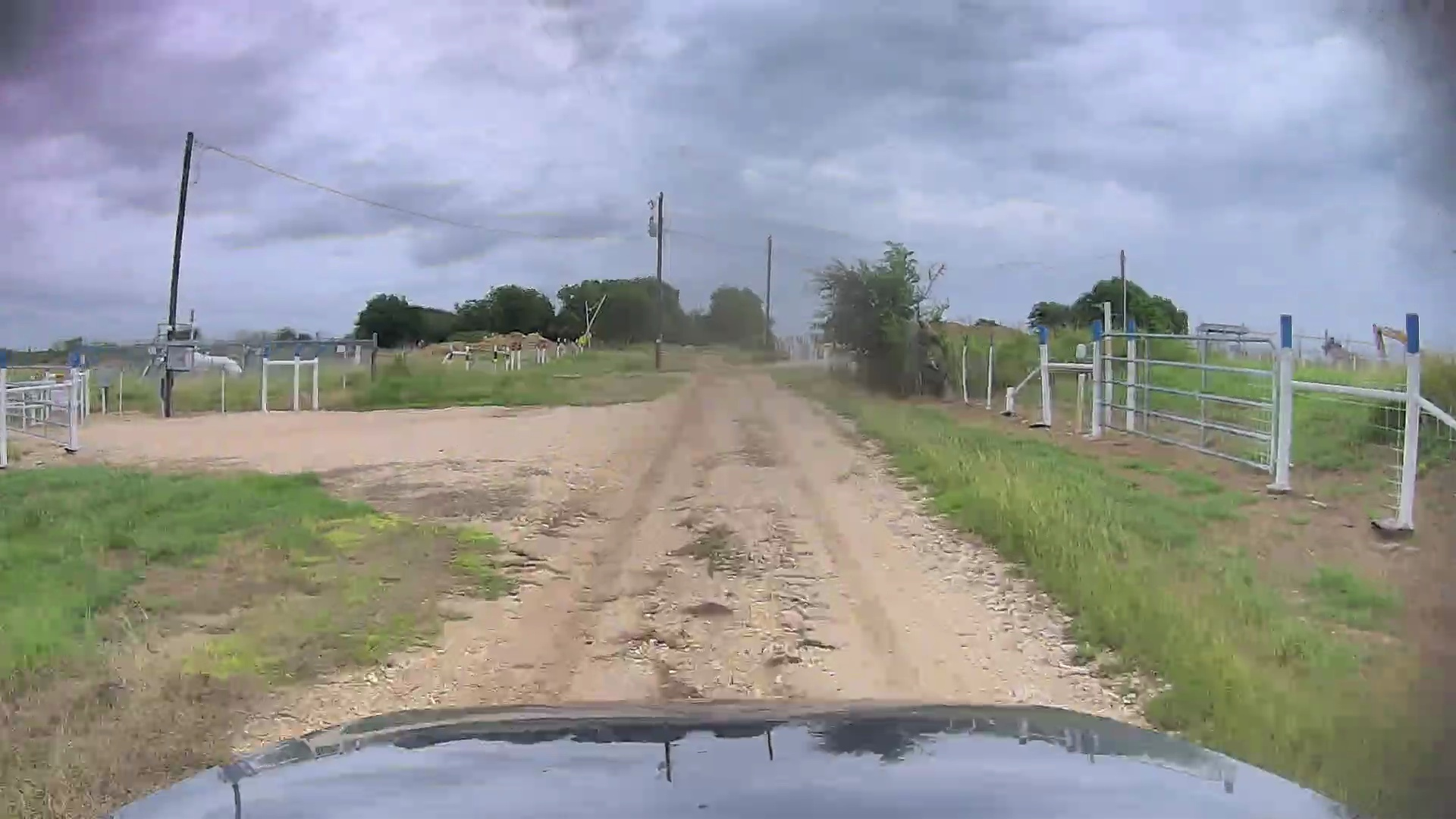} &
        \includegraphics[width=0.22\textwidth]{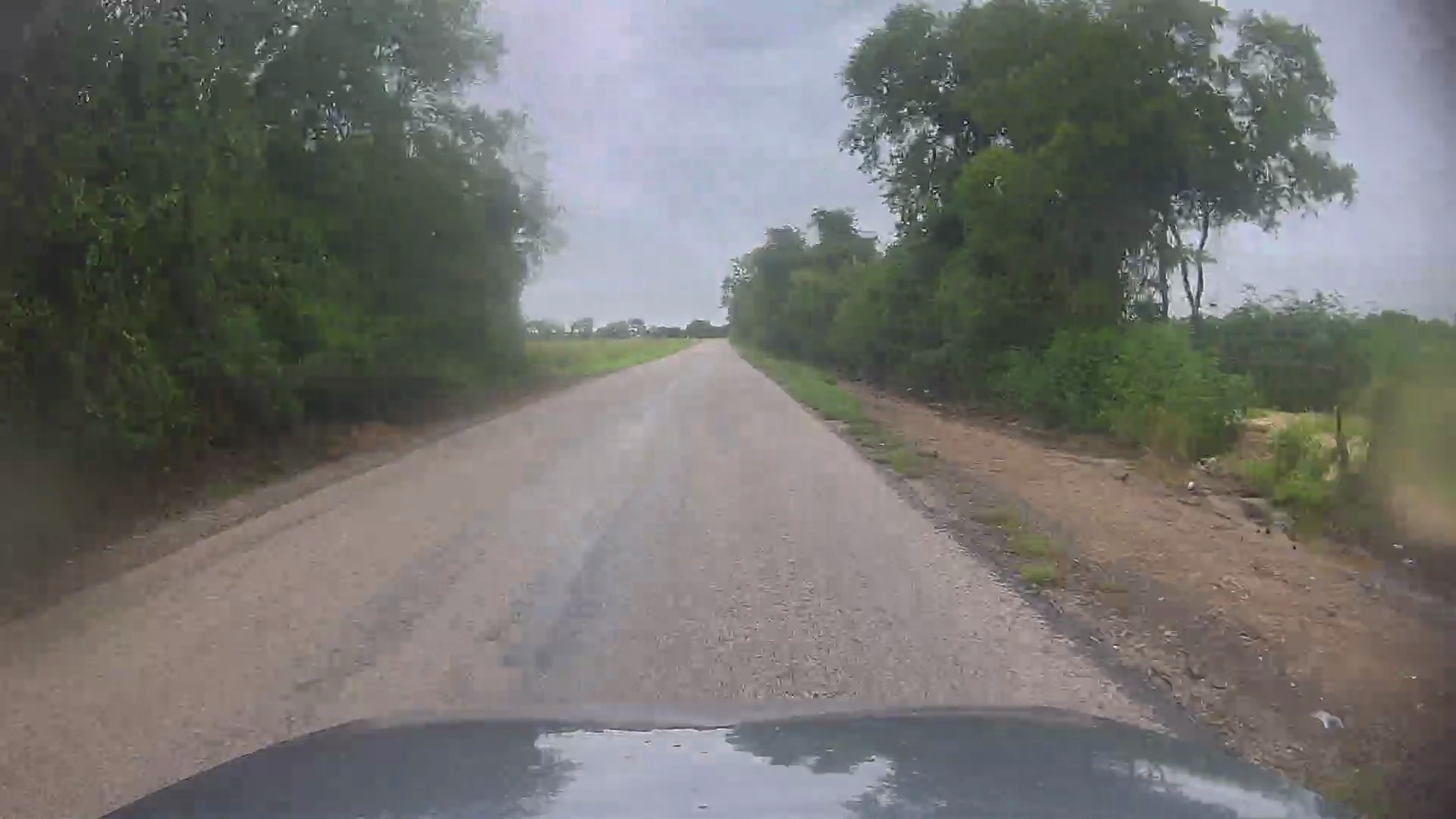} \\

        \includegraphics[width=0.22\textwidth]{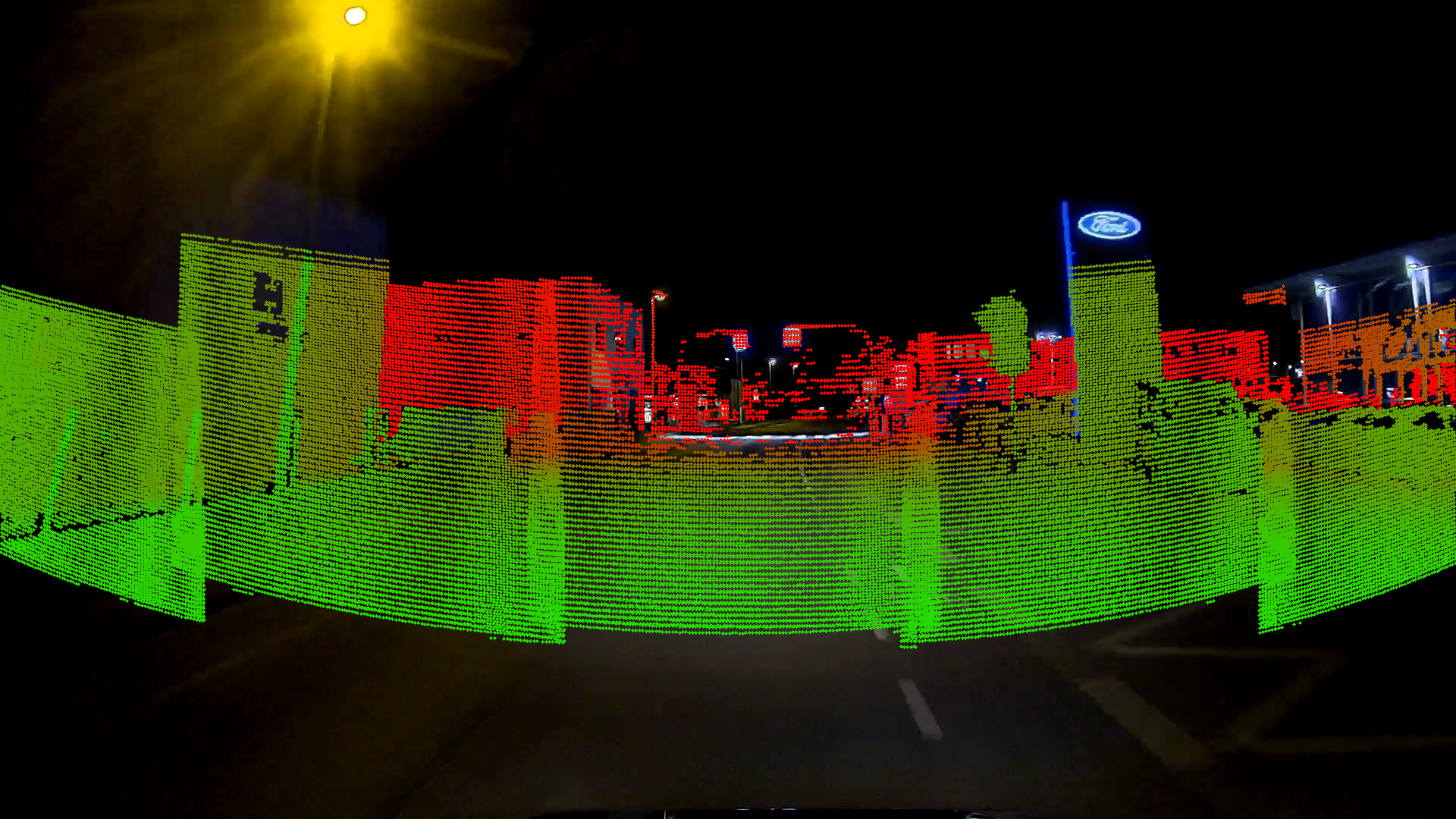} &
        \includegraphics[width=0.22\textwidth]{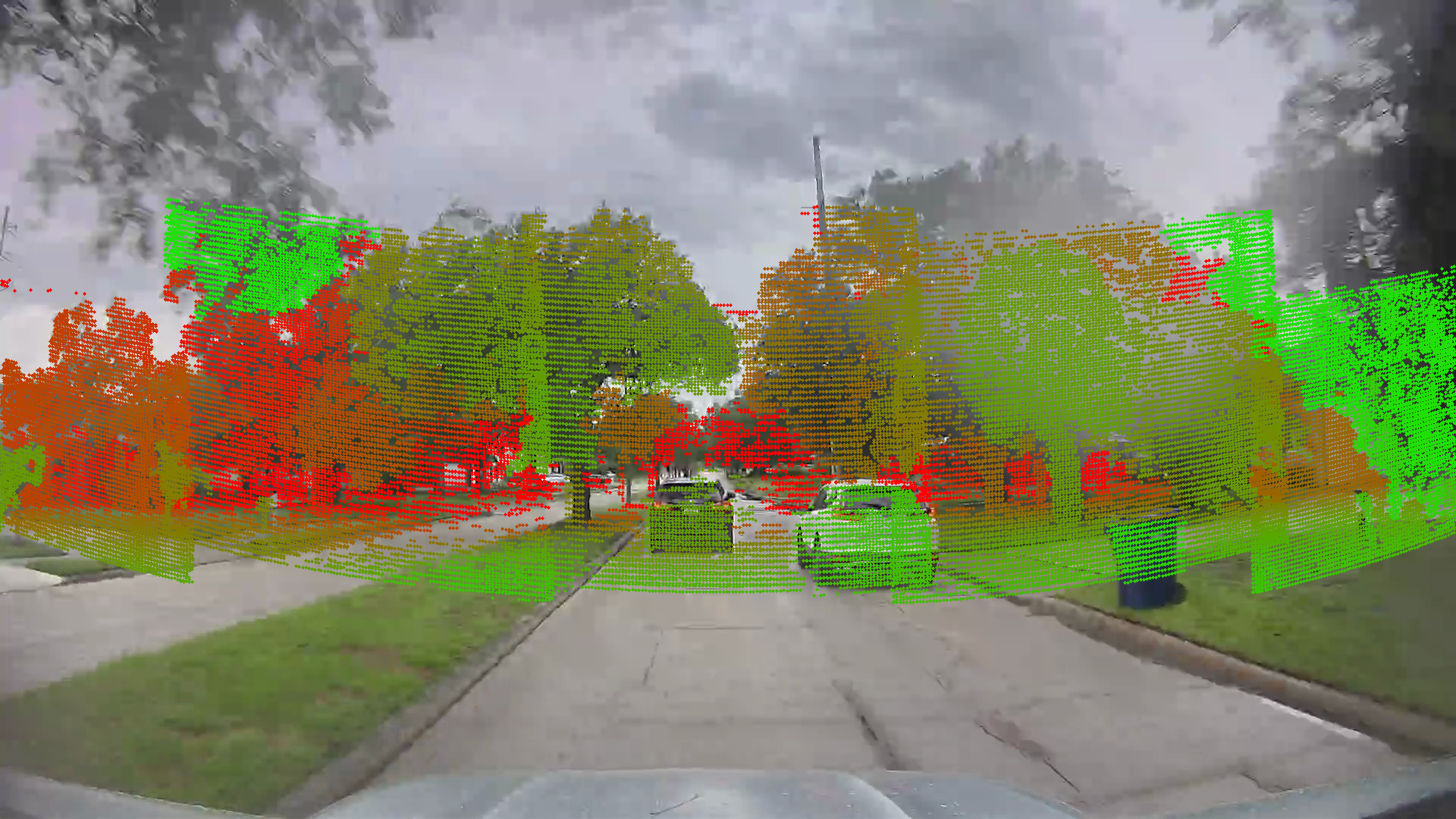} &
        \includegraphics[width=0.22\textwidth]{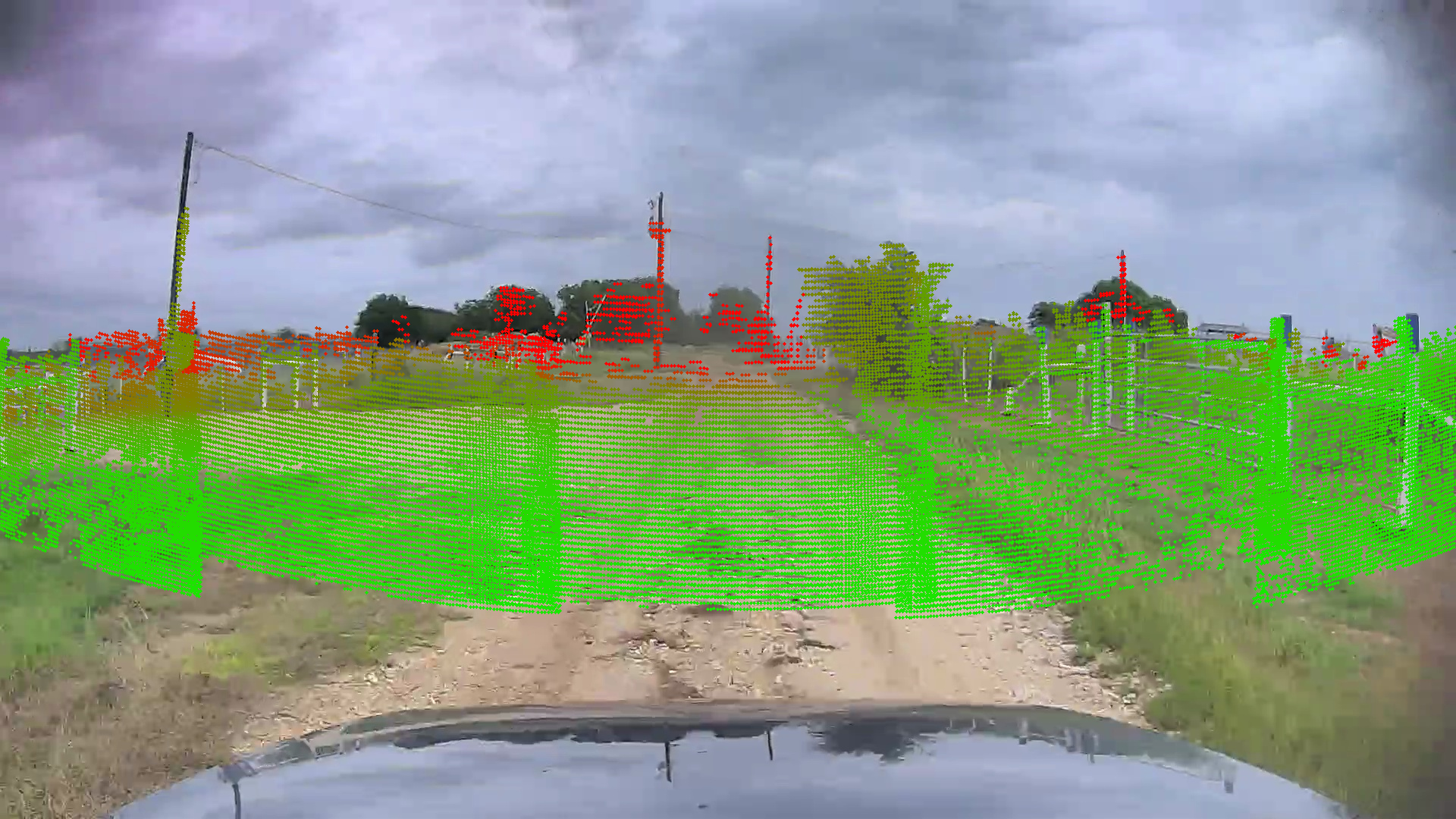} &
        \includegraphics[width=0.22\textwidth]{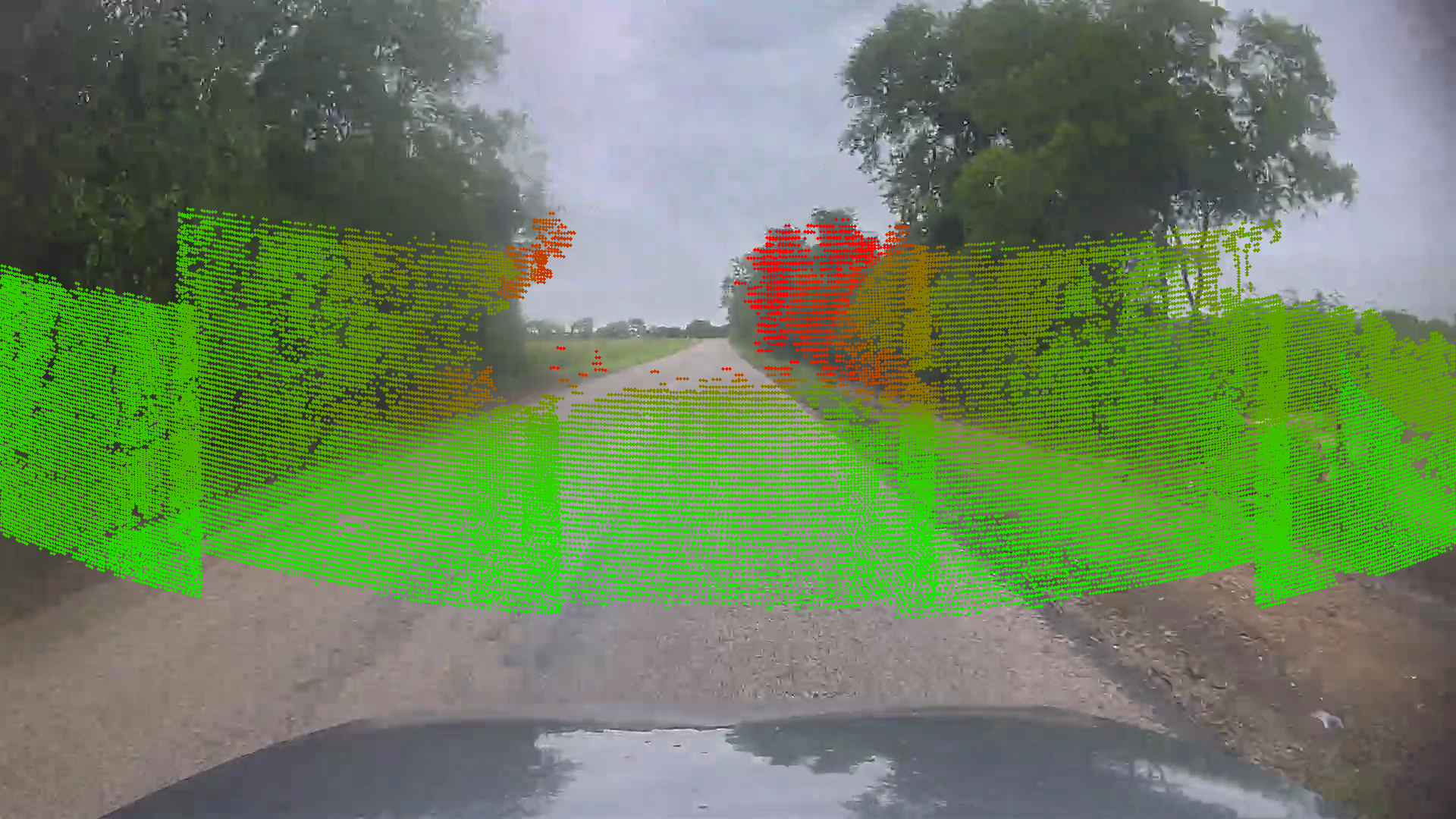} \\

        (e) Urban (Night) & (f) Urban (Rainy) & (g) Rural (Normal) & (h) Rural (Rainy)
    \end{tabular}

    \vspace{-1mm}
    \captionof{figure}{\textbf{Data visualization of the ROVR dataset.} The first and third rows show RGB images, while the second and fourth rows present the corresponding depth maps projected onto the RGB images.}
    \label{fig:stereocarla}
    \vspace{-3mm}
\end{minipage}

\subsection{Dataset Statistics}
\label{sec:statistics}
Table~\ref{tab:stereo_data} summarizes the frame counts across scene types and conditions. The training split contains 193{,}648 frames and the test split 10{,}002 frames, giving a 95.1\% / 4.9\% train/test ratio. The condition distribution follows naturalistic driving frequency: highway-normal is the majority ($\sim$70\% of the training split), reflecting real-world fleet operation, while night ($\sim$11\%) and rain ($\sim$4\%) are under-represented but deliberately preserved as \emph{stress-test} splits. We intentionally do \emph{not} rebalance these minority splits because one of the goals of ROVR is to characterize how current models fail when the operating condition is rare, rather than to enforce artificial parity.

\paragraph{Depth range and density.}
Valid LiDAR returns in the training split span $1$--$200\,\mathrm{m}$, with a median of $18.2\,\mathrm{m}$ and the 95th percentile at $96\,\mathrm{m}$. For training and evaluation we clip depths to $[1, 80]\,\mathrm{m}$ following common practice. Across all frames, the mean valid-pixel density after clipping is $3.4\%$ of image pixels on highway, $2.7\%$ on urban, and $2.1\%$ on rural---substantially sparser than the dense training labels of DDAD, and comparable to projected KITTI depth. The anisotropy noted in Sec.~\ref{sec:sensor-suite} manifests here as a density peak between elevations $-3^{\circ}$ and $+1^{\circ}$, covering the road surface and nearby vehicles, and a long sparse tail for elevated structures.

\paragraph{Geographic coverage.}
The fleet has recorded in North America, Europe, and Asia, with the current release drawing from metropolitan regions on each continent. The sequence metadata includes a coarse region tag so that users can partition the data geographically for out-of-region generalization studies. Although the full mileage exceeds 10{,}000 hours, this release deliberately restricts itself to a depth-estimation subset with verified calibration and privacy blurring.

\section{Experiments}
\label{sec:experiments}

We conduct experiments on the ROVR dataset to establish baseline performance and to analyse the influence of the key environmental factors that distinguish ROVR from prior benchmarks. The evaluation is organized along three axes. First, we report benchmark results on the full test split, providing a reference point that future studies can use directly. Second, we run two cross-dataset studies---existing benchmark $\to$ ROVR and ROVR $\to$ existing benchmarks---to quantify generalization in both directions. Third, we conduct domain-specific ablations under varying illumination and scene types to characterize \emph{when} and \emph{why} current methods fail. Throughout, we train every model with identical pre-processing, optimization, and clipping ranges so that performance differences cleanly reflect model capacity rather than experimental luck.

\subsection{Implementation Details}
We evaluate two families of depth estimation baselines.

\textbf{Zero-Shot Models.} These models are trained on web-scale data to predict either affine-invariant or metric depth, targeting zero-shot generalization to unseen domains. We select two affine-invariant baselines---Marigold v1.1~\cite{marigold} and DepthAnythingV2~\cite{depthanythingv2}---and four metric baselines: KITTI-finetuned DepthAnythingV2~\cite{depthanythingv2}, Depth Pro~\cite{Bochkovskii2024:arxiv}, UniK3D~\cite{piccinelli2025unik3d}, and UniDepth v2~\cite{piccinelli2025unidepthv2}. We use the official weights provided by each author, selecting the variant yielding the best results on the validation subset. For affine-invariant models we apply median-scale alignment per-image before computing metric errors, which is the standard protocol in the monocular depth literature.

\textbf{Video Depth Models.} To assess whether temporal information alleviates the challenges posed by ROVR, we additionally evaluate one video-based depth estimation method. Temporal methods can in principle exploit parallax from ego motion to disambiguate scale, so they provide an orthogonal test of the difficulty of ROVR.

\textbf{Single-Domain Models.} These models are trained to estimate metric depth on a single large-scale dataset, aiming to excel within a specific domain. We select three representative baselines: VA-DepthNet~\cite{va-depthnet}, DCDepth~\cite{dcdepth}, and IEBins~\cite{iebins}. To ensure comparability across methods, we reproduce the variants that adopt the Swin Transformer–Large (Swin-L)~\cite{swin} backbone and initialize training solely from its pretrained weights---no auxiliary self-supervised or cross-domain pretraining. All images are center-cropped to $960\times 544$ for both training and evaluation, and each model is trained for five epochs. Unless otherwise noted, we follow the official training configurations of each method (e.g., learning rate, weight decay, optimizer schedule). Ground-truth depth is measured in metres; during training and evaluation, depths are clipped to $[1, 80]$ metres and invalid pixels are masked. All experiments are conducted on $8\times$ NVIDIA L20 GPUs with mixed-precision training, which keeps the per-model wall-clock under 48 hours and makes the full benchmark reproducible within a one-week compute budget.

For KITTI experiments, we follow the \textit{kitti\_eigen} preprocessing: images are resized to $352\times1216$ for both training and evaluation. For DDAD, nuScenes, and ROVR, images are center-cropped to $960\times544$. In cross-dataset evaluation, KITTI results use the official pretrained weights, while evaluation on other datasets applies center cropping followed by resizing to the training resolution. We deliberately avoid per-dataset tuning of crop size or learning rate, since the goal of the cross-dataset study is to characterize out-of-distribution behaviour, not to extract the best possible number on each benchmark.

\subsection{Evaluation Metrics}
We follow standard monocular depth estimation~\cite{zhao2022monovit, duan2023diffusiondepth} practice and report six error metrics where lower is better: scale-invariant log RMSE (silog), absolute relative error (abs\_rel), log10 error (log10), root mean squared error (rms), squared relative error (sq\_rel), and log RMSE (log\_rms). We also report three accuracy metrics where higher is better: $\delta_1$, $\delta_2$, and $\delta_3$, representing the fraction of pixels where the prediction is within a factor of $1.25$, $1.25^2$, and $1.25^3$ of the ground truth. The three $\delta$ thresholds together summarize the shape of the error distribution: $\delta_1$ captures ``near-perfect'' predictions, $\delta_2$ captures useful predictions that are close enough for most downstream tasks, and $\delta_3$ captures the tail of catastrophic failures. When only one number is reported in the text, we use abs\_rel and $\delta_1$ as the primary accuracy indicators, consistent with the majority of prior work.

\begin{table}[t]
  \centering
  \caption{\textbf{Performance of depth estimation models across KITTI, DDAD, nuScenes, and ROVR.} Models trained on existing datasets perform well on their own test sets but fail to generalize to ROVR.
  }
  \label{tab:k_d_n_Rovr}
  \fontsize{8pt}{9.5pt}\selectfont
  \setlength\tabcolsep{1pt}
  \renewcommand\arraystretch{1.05}
  \begin{tabular}{lc|cccccc|ccc}
    \rowcolor{gray!20}
    \hline
    && \multicolumn{6}{c|}{\textbf{lower is better $\downarrow$}} & \multicolumn{3}{c}{\textbf{higher is better $\uparrow$}} \\
    \rowcolor{gray!20}
    \multirow{-2}{*}{\textbf{Train}}&\multirow{-2}{*}{\textbf{Test}}& silog & abs\_rel & log10 & rms & sq\_rel & log\_rms &$\delta_1$ & $\delta_2$ & $\delta_3$  \\
    
    \hline
    \multicolumn{10}{l}{\textbf{VA-DepthNet~\cite{va-depthnet}}} \\ 
    \midrule
     KITTI~\cite{kittidepth}&KITTI~\cite{kittidepth}& 6.71 & 0.05 & 0.02 & 2.02 & 0.14 & 0.07 & 0.98 & 1.00 & 1.00\\
     \rowcolor{gray!10}
    KITTI~\cite{kittidepth}&ROVR (Ours)& 49.65 & 0.63 & 0.30 & 14.91 & 11.80 & 0.76 & 0.09 & 0.22 & 0.50\\
    \midrule
    DDAD~\cite{Guizilini2020ddad}&DDAD~\cite{Guizilini2020ddad}& 11.76 & 0.09 & 0.04 & 5.06 & 0.74 & 0.13 & 0.91 & 0.98 & 0.99\\
    \rowcolor{gray!10}
    DDAD~\cite{Guizilini2020ddad}&ROVR (Ours)& 37.13 & 0.75 & 0.18 & 13.16 & 20.76 & 0.54 & 0.23 & 0.67 & 0.88\\
    \midrule
    nuScenes~\cite{nuscenes}&nuScenes~\cite{nuscenes}& 8.25 & 0.05 & 0.02 & 2.28 & 0.36 & 0.09 & 0.97 & 0.98 & 0.99\\
    \rowcolor{gray!10}
    nuScenes~\cite{nuscenes}&ROVR (Ours)& 50.75 & 0.49 & 0.16 & 12.59 & 10.49 & 0.52 & 0.42 & 0.70 & 0.84\\
     
    \bottomrule
  \end{tabular}%
\end{table}

\begin{table}[t]
  \centering
  \caption{\textbf{Cross-dataset evaluation} of three depth estimation models: training on KITTI leads to large performance drops on ROVR, while training on ROVR yields better results.
  }
  \label{tab:kitti_rovr}
  \fontsize{8pt}{9.5pt}\selectfont
  \setlength\tabcolsep{1pt}
  \renewcommand\arraystretch{1.05}
  \begin{tabular}{lc|cccccc|ccc}
    \hline
    \rowcolor{gray!20}
    && \multicolumn{6}{c|}{\textbf{lower is better $\downarrow$}} & \multicolumn{3}{c}{\textbf{higher is better $\uparrow$}} \\ 
    \rowcolor{gray!20}
     \multirow{-2}{*}{\textbf{Train}}&\multirow{-2}{*}{\textbf{Test}}& silog & abs\_rel & log10 & rms & sq\_rel & log\_rms &$\delta_1$ & $\delta_2$ & $\delta_3$  \\
    
    \hline
    \multicolumn{10}{l}{\textbf{VA-DepthNet~\cite{va-depthnet}}} \\ \midrule
     KITTI~\cite{kittidepth}&ROVR (Ours)& 48.76 & 0.59 & 0.24 & 13.78 & 12.27 & 0.64 & 0.15 & 0.42 & 0.75\\
     \rowcolor{gray!10}
    ROVR (Ours)&ROVR (Ours)&29.33 & 0.18 & 0.05 & 6.95 & 3.95 & 0.30 & 0.89 & 0.94 & 0.96 \\ 
    \midrule
    \multicolumn{10}{l}{\textbf{DCDepth~\cite{dcdepth}}} \\ \midrule
     KITTI~\cite{kittidepth}&ROVR (Ours)& 49.93 & 0.64 & 0.29 & 14.98 & 12.41 & 0.74 & 0.07 & 0.21 & 0.59\\
     \rowcolor{gray!10}
    ROVR (Ours)&ROVR (Ours)& 27.28 & 0.19 & 0.05 & 6.46 & 4.19 & 0.28 & 0.89 & 0.94 & 0.96 \\
    
    \midrule\midrule
    \multicolumn{10}{l}{\textbf{IEBins~\cite{iebins}}} \\ \midrule
    KITTI~\cite{kittidepth}&ROVR (Ours)& 48.50 & 0.61 & 0.30 & 15.22 & 10.76 & 0.76 & 0.07 & 0.20 & 0.58\\
     \rowcolor{gray!10}
    ROVR (Ours)&ROVR (Ours)& 29.97 & 0.21 & 0.07 & 7.37 & 3.75 &  0.31 & 0.84 & 0.92 & 0.95 \\
    
    \bottomrule
  \end{tabular}%
\end{table}

\subsection{Main Results}

As shown in Table~\ref{tab:k_d_n_Rovr}, models trained on KITTI~\cite{kittidepth}, DDAD~\cite{Guizilini2020ddad}, or nuScenes~\cite{nuscenes} achieve strong in-domain results but suffer severe degradation when evaluated on ROVR, revealing poor cross-dataset generalization. In particular, abs\_rel increases by an order of magnitude and $\delta_1$ drops from $>0.98$ on KITTI to $<0.25$ on ROVR. The degradation is consistent across architectures: VA-DepthNet, DCDepth, and IEBins, which use very different inductive biases, all collapse to the same $\delta_1$ regime, indicating that the gap is primarily due to the distribution shift of ROVR rather than to any one model's weakness.

These results expose two key issues. First, existing benchmarks are \emph{saturated}: state-of-the-art methods achieve near-ceiling scores on KITTI, yet the same models fail on ROVR, demonstrating that high in-domain accuracy does not imply robustness. Second, even when trained directly on ROVR, current methods remain far from saturation (best abs\_rel $\sim 0.18$--$0.21$, Table~\ref{tab:kitti_rovr}), confirming that ROVR presents a substantially more challenging setting. These findings motivate the development of new depth estimation approaches capable of handling greater scene diversity and sparser ground truth.

A useful interpretation of the cross-dataset drop is that KITTI-trained models have learned a \emph{dataset-specific} prior over road geometry, ego-motion, and camera intrinsics---an implicit prior that happens to be well-specified for the Karlsruhe recording rig but does not transfer. When that prior collides with ROVR's broader distribution (three continents, three climate regimes, three scene types), predictions collapse. This is not surprising, but the magnitude of the collapse is: the same models that score $\delta_1\!>\!0.98$ in-domain regress to $\delta_1\!<\!0.25$ out-of-domain, a range that is effectively unusable for any downstream planner. It is this gap, not the raw in-domain accuracy, that should be the focus of future benchmarking in autonomous driving.

\textbf{Reverse cross-dataset evaluation.}
To verify that ROVR training also benefits generalization \emph{to} existing benchmarks, we evaluate ROVR-trained models on KITTI, nuScenes, and DDAD test sets. A dataset that is strictly harder than its peers should, in the limit, produce models that transfer better to easier settings---so the reverse transfer is a direct stress test of the generality of ROVR-trained features.

\subsection{Ablation Study}
Tables~\ref{tab:resillum} and~\ref{tab:scenetypes} present depth estimation performance across illumination conditions and scene types, respectively. We report results for both zero-shot and single-domain models to provide a comprehensive analysis, and we discuss each factor in turn.

Notably, the zero-shot metric models produce near-identical scores across conditions. This is because their predictions, while visually plausible, are scale-misaligned with the ROVR ground truth---the affine or metric calibration learned from other domains does not transfer, collapsing their effective accuracy to a similar failure mode regardless of the input condition. In contrast, single-domain models trained on ROVR exhibit meaningful variation across conditions, making them more informative for ablation analysis. We therefore focus the discussion below on the ROVR-trained single-domain models, returning to the zero-shot baselines only when a contrast is informative.

\textbf{Results under varying illumination.} Table~\ref{tab:resillum} breaks down performance by lighting condition (Normal, Night, Rainy). All methods perform best under normal lighting, as expected. Performance degrades noticeably at night, with $\delta_1$ and $\delta_2$ declining across the board; VA-DepthNet is the most robust in this setting, achieving the lowest abs\_rel and rms. The largest degradation occurs under rainy conditions, where both RGB quality and LiDAR density are adversely affected---wet asphalt exhibits specular reflection that returns fewer valid LiDAR points, and raindrops on the windshield introduce local photometric noise that disturbs the monocular cues the model relies on. IEBins is the most affected, exhibiting substantially higher log10 and sq\_rel, suggesting that its bin-based depth head is particularly sensitive to the compounded challenges of image noise and sparse supervision.

\begin{table}[!t]
\centering
\caption{\textbf{Depth estimation performance under varying illumination (\emph{light conditions})}. Zero-shot methods are divided into affine-invariant (top) and metric (bottom).}
\fontsize{8pt}{9.5pt}\selectfont
\setlength\tabcolsep{1.5pt}
\renewcommand\arraystretch{1.08}
\begin{tabular*}{\linewidth}{@{\extracolsep{\fill}}c|c|c||cccccc|ccc@{}}
\hline
\rowcolor{gray!20}
 &  && \multicolumn{6}{c|}{\textbf{Lower is better $\downarrow$}} & \multicolumn{3}{c}{\textbf{Higher is better$\uparrow$}} \\ 
\rowcolor{gray!20}
\multirow{-2}{*}{\textbf{Domain}} & \multirow{-2}{*}{\makecell{\textbf{Model}\\\textbf{Type}}} &\multirow{-2}{*}{\textbf{Method}} & silog & abs\_rel & log10 & rms & sq\_rel & log\_rms & $\delta_1$ & $\delta_2$ & $\delta_3$ \\ 
\midrule

\multirow{9}{*}{\textbf{Normal}}
&&Marigold v1.1~\cite{marigold}&56.93 &1.31 &0.15 &11.27 &48.04 &0.58 &0.54 &0.79 &0.88\\
&&DepthAnythingV2~\cite{depthanythingv2}&117.90 &1.17 &0.41 &21.79 &28.62 &1.20 &0.15 &0.29 &0.43\\
\cline{3-12}
        &&DepthAnythingV2~\cite{depthanythingv2}
        &52.34 &3.98 &0.50 &37.63 &198.82 &1.28 &0.05 &0.11 &0.18\\
        &&Depth Pro \cite{Bochkovskii2024:arxiv}&51.46 &0.99 &0.38 &15.88 &20.39 &0.98 &0.11 &0.28 &0.47\\
        
        &&UniK3D~\cite{piccinelli2025unik3d}&56.41 &1.81 &0.16 &13.92 &91.57 &0.61 &0.44 &0.83 &0.90\\
        &\multirow{-6}{*}{\makecell{\textbf{Zero-}\\\textbf{Shot}}}&UniDepth v2 \cite{piccinelli2025unidepthv2}&55.19 &1.76 &0.13 &13.54 &91.06 &0.58 &0.65 &0.87 &0.91\\ 
        \cline{2-12}
        && VA-DepthNet~\cite{va-depthnet} & 29.89 & 0.18 & 0.05 & 7.04 & 4.09 & 0.30 & 0.89 & 0.94 & 0.96  \\
        && DCDepth~\cite{dcdepth} &  29.67 & 0.20 & 0.06 & 7.03 & 4.37 & 0.30 & 0.88 & 0.93 & 0.96 \\
        &\multirow{-3}{*}{\makecell{\textbf{Single-}\\\textbf{Domain}}}& IEBins~\cite{iebins} & 30.38 & 0.21 & 0.06 & 7.45 & 3.78 & 0.31 & 0.85 & 0.92 & 0.95\\
        \midrule
\multirow{9}{*}{\textbf{Night}} &&Marigold v1.1~\cite{marigold}&46.97 &0.62 &0.15 &9.78 &13.33 &0.49 &0.47 &0.75 &0.88\\
&&DepthAnythingV2~\cite{depthanythingv2}&109.67 &0.82 &0.37 &16.92 &13.68 &1.13 &0.17 &0.34 &0.50\\
\cline{3-12}
        &&DepthAnythingV2~\cite{depthanythingv2}
        &51.75 &2.49 &0.43 &27.84 &92.14 &1.08 &0.07 &0.15 &0.23\\
        &&Depth Pro \cite{Bochkovskii2024:arxiv}&53.33 &0.72 &0.37 &14.34 &13.96 &1.00 &0.17 &0.34 &0.49\\
        
        &&UniK3D~\cite{piccinelli2025unik3d}&42.95 &0.68 &0.14 &10.09 &19.86 &0.49 &0.52 &0.79 &0.88\\
        &\multirow{-6}{*}{\makecell{\textbf{Zero-}\\\textbf{Shot}}}&UniDepth v2 \cite{piccinelli2025unidepthv2}&43.21 &0.68 &0.12 &10.22 &22.09 &0.46 &0.64 &0.82 &0.89\\ 
        \cline{2-12}
        & &  VA-DepthNet~\cite{va-depthnet} & 27.37 & 0.17 & 0.05 & 6.68 & 3.43 & 0.28 & 0.88 & 0.94 & 0.96  \\
        && DCDepth~\cite{dcdepth} &  27.24 & 0.18 & 0.05 & 6.56 & 3.57 & 0.28 & 0.87 & 0.94 & 0.96 \\
        &\multirow{-3}{*}{\makecell{\textbf{Single-}\\\textbf{Domain}}}& IEBins~\cite{iebins} & 28.34 & 0.21 & 0.06 & 6.84 & 3.79 & 0.29 & 0.85 & 0.93 & 0.96\\
         \midrule
\multirow{9}{*}{\textbf{Rainy}}  
&&Marigold v1.1~\cite{marigold}&45.11 &0.78 &0.13 &8.77 &24.12 &0.46 &0.58 &0.82 &0.91\\
        &&DepthAnythingV2~\cite{depthanythingv2}&115.65 &1.00 &0.40 &18.05 &19.85 &1.18 &0.16 &0.31 &0.45\\
        \cline{3-12}
        &&DepthAnythingV2~\cite{depthanythingv2}
        &44.86 &3.70 &0.56 &37.42 &162.81 &1.37 &0.03 &0.07 &0.13\\
        &&Depth Pro \cite{Bochkovskii2024:arxiv}&46.15 &0.75 &0.48 &15.68 &10.84 &1.18 &0.02 &0.09 &0.25\\
        &&UniK3D~\cite{piccinelli2025unik3d}&45.29 &0.97 &0.12 &10.25 &42.89 &0.47 &0.59 &0.82 &0.92\\
        &\multirow{-6}{*}{\makecell{\textbf{Zero-}\\\textbf{Shot}}}&UniDepth v2 \cite{piccinelli2025unidepthv2}&43.02 &0.98 &0.12 &10.05 &42.90 &0.45 &0.60 &0.88 &0.94\\ 
        \cline{2-12}
        &&  VA-DepthNet~\cite{va-depthnet} & 25.52 & 0.17 & 0.05 & 6.34 & 3.08 & 0.26 & 0.87 & 0.94 & 0.96 \\
        && DCDepth~\cite{dcdepth} &  25.28 & 0.18 & 0.05 & 6.27 & 3.02 & 0.26 & 0.86 & 0.93 & 0.96 \\
        &\multirow{-3}{*}{\makecell{\textbf{Single-}\\\textbf{Domain}}}& IEBins~\cite{iebins} & 27.65 & 0.29 & 0.10 & 7.32 & 3.24 & 0.32 & 0.70 & 0.92 & 0.95 \\
\bottomrule
\end{tabular*}

\label{tab:resillum}
\end{table}

A secondary observation is that the relative ordering of methods changes with the condition: IEBins outperforms DCDepth in Normal but underperforms it in Rainy, implying that there is no single architecture that dominates. This makes ROVR useful as a diagnostic benchmark, since it rewards approaches that are robust across conditions rather than peak approaches tuned to one regime.

\begin{table}[!t]
\centering
\caption{\textbf{Depth estimation performance across scene types}. Zero-shot methods are divided into affine-invariant (top) and metric (bottom).}
\fontsize{8pt}{9.5pt}\selectfont
\setlength\tabcolsep{1.4pt}
\renewcommand\arraystretch{1.08}
\begin{tabular*}{\linewidth}{@{\extracolsep{\fill}}c|c|c||cccccc|ccc@{}}
\hline
\rowcolor{gray!20}
 &  && \multicolumn{6}{c|}{\textbf{Lower is better $\downarrow$}} & \multicolumn{3}{c}{\textbf{Higher is better$\uparrow$}} \\ 
\rowcolor{gray!20}
\multirow{-2}{*}{\textbf{Domain}} & \multirow{-2}{*}{\makecell{\textbf{Model}\\\textbf{Type}}} &\multirow{-2}{*}{\textbf{Method}} & silog & abs\_rel & log10 & rms & sq\_rel & log\_rms & $\delta_1$ & $\delta_2$ & $\delta_3$ \\ 
\midrule

\multirow{9}{*}{\textbf{Highway}} 
        &&Marigold v1.1~\cite{marigold}&55.23 &1.21 &0.15 &11.10 &43.68 &0.56 &0.54 &0.79 &0.88\\
        &&DepthAnythingV2~\cite{depthanythingv2}&117.95 &1.08 &0.41 &20.50 &23.92 &1.21 &0.15 &0.30 &0.44\\
        \cline{3-12}
        &&DepthAnythingV2~\cite{depthanythingv2}
        &51.80 &3.76 &0.49 &35.43 &179.93 &1.25 &0.05 &0.11 &0.18\\
        &&Depth Pro \cite{Bochkovskii2024:arxiv}&51.75 &0.95 &0.36 &15.16 &19.05 &0.94 &0.11 &0.28 &0.48\\

        &&UniK3D~\cite{piccinelli2025unik3d}&54.86 &1.64 &0.16 &13.52 &80.15 &0.60 &0.42 &0.82 &0.90\\
        &\multirow{-6}{*}{\makecell{\textbf{Zero-}\\\textbf{Shot}}}&UniDepth v2 \cite{piccinelli2025unidepthv2}&53.65 &1.61 &0.13 &13.24 &81.73 &0.57 &0.64 &0.86 &0.91\\ 
        \cline{2-12}
        & &  VA-DepthNet~\cite{va-depthnet} & 26.28 & 0.15 & 0.04 & 6.22 & 3.12 & 0.27 & 0.91 & 0.95 & 0.97 \\
        && DCDepth~\cite{dcdepth} &  26.36 & 0.16 & 0.04 & 6.27 & 3.31 & 0.27 & 0.90 & 0.95 & 0.97 \\
        &\multirow{-3}{*}{\makecell{\textbf{Single-}\\\textbf{Domain}}}& IEBins~\cite{iebins} & 27.09 & 0.18 & 0.06 & 6.62 & 3.03 & 0.28 & 0.87 & 0.94 & 0.96 \\
         \midrule
\multirow{9}{*}{\textbf{Rural}}          &&Marigold v1.1~\cite{marigold}&47.22 &0.97 &0.12 &9.54 &35.04 &0.48 &0.59 &0.85 &0.93\\

&&DepthAnythingV2~\cite{depthanythingv2}&112.14 &0.99 &0.39 &19.58 &20.82 &1.14 &0.16 &0.31 &0.45\\
\cline{3-12}
        &&DepthAnythingV2~\cite{depthanythingv2}
        &47.84 &3.56 &0.52 &37.91 &166.56 &1.29 &0.04 &0.09 &0.16\\
        &&Depth Pro \cite{Bochkovskii2024:arxiv}&48.01 &0.85 &0.34 &14.33 &16.35 &0.89 &0.11 &0.28 &0.49\\
        &&UniK3D~\cite{piccinelli2025unik3d}&49.24 &1.14 &0.15 &11.51 &49.37 &0.53 &0.42 &0.80 &0.91\\
        &\multirow{-6}{*}{\makecell{\textbf{Zero-}\\\textbf{Shot}}}&UniDepth v2 \cite{piccinelli2025unidepthv2}&47.30 &1.14 &0.12 &11.21 &53.43 &0.49 &0.63 &0.87 &0.93\\ 
        \cline{2-12}
        &&  VA-DepthNet~\cite{va-depthnet} & 29.17 & 0.18 & 0.05 & 6.91 & 4.14 & 0.29 & 0.90 & 0.94 & 0.96 \\
        && DCDepth~\cite{dcdepth} &  29.22 & 0.20 & 0.05 & 6.79 & 4.64 & 0.29 & 0.89 & 0.94 & 0.96 \\
        &\multirow{-3}{*}{\makecell{\textbf{Single-}\\\textbf{Domain}}}& IEBins~\cite{iebins} & 29.93 & 0.24 & 0.07 & 7.47 & 3.89 & 0.32 & 0.84 & 0.93 & 0.95 \\
         \midrule
\multirow{9}{*}{\textbf{Urban}}   &&Marigold v1.1~\cite{marigold}&57.47 &1.25 &0.16 &11.05 &41.91 &0.59 &0.49 &0.76 &0.87\\
        &&DepthAnythingV2~\cite{depthanythingv2}&115.24 &1.28 &0.41 &22.90 &35.16 &1.17 &0.16 &0.31 &0.44\\
        \cline{3-12}
        &&DepthAnythingV2~\cite{depthanythingv2}
        &53.40 &3.97 &0.50 &38.79 &201.98 &1.28 &0.06 &0.12 &0.19\\
        &&Depth Pro \cite{Bochkovskii2024:arxiv}&51.54 &0.97 &0.48 &17.67 &20.09 &1.19 &0.11 &0.25 &0.38\\
        &&UniK3D~\cite{piccinelli2025unik3d}&53.95 &1.78 &0.14 &13.12 &91.66 &0.58 &0.57 &0.84 &0.91\\
        &\multirow{-6}{*}{\makecell{\textbf{Zero-}\\\textbf{Shot}}}&UniDepth v2 \cite{piccinelli2025unidepthv2}&53.48 &1.69 &0.13 &12.73 &85.04 &0.56 &0.66 &0.86 &0.91\\ 
        \cline{2-12}
        &&  VA-DepthNet~\cite{va-depthnet} & 38.01 & 0.28 & 0.08 & 9.05 & 6.23 & 0.39 & 0.81 & 0.90 & 0.93 \\
        && DCDepth~\cite{dcdepth} &  36.88 & 0.30 & 0.08 & 8.85 & 6.55 & 0.38 & 0.79 & 0.90 & 0.94 \\
        &\multirow{-3}{*}{\makecell{\textbf{Single-}\\\textbf{Domain}}}& IEBins~\cite{iebins} & 38.12 & 0.30 & 0.09 & 9.45 & 5.73 & 0.39 & 0.76 & 0.88 & 0.92 \\
\bottomrule
\end{tabular*}

\label{tab:scenetypes}
\end{table}

\textbf{Results across scene types.} Table~\ref{tab:scenetypes} stratifies performance by scene type (Highway, Rural, Urban). Highway scenes yield the best results for all methods, consistent with their simpler geometry and more uniform lighting. Urban scenes are the most challenging: dense clutter, frequent occlusions, and multi-scale objects lead to markedly higher errors across all metrics, with IEBins showing the largest gap. These results confirm that scene complexity---not just data scale---is a primary driver of depth estimation difficulty. Rural scenes sit between highway and urban, and they expose a failure mode that is specific to ROVR: the combination of foliage clutter, weak photometric cues on unpaved surfaces, and long open stretches of sky creates a bimodal depth distribution that is harder for bin-based predictors than the roughly unimodal KITTI distribution.

\begin{wrapfigure}[10]{r}{0.42\linewidth}
    \centering
    \vspace{-2mm}
    \includegraphics[width=\linewidth]{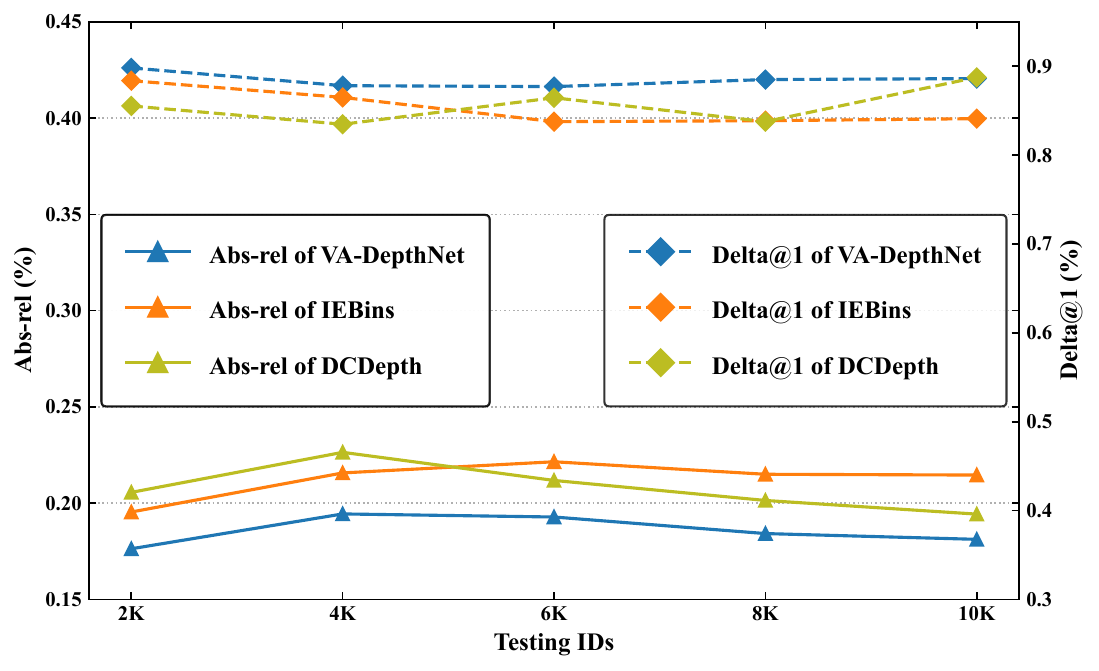}
    \caption{\textbf{Abs\_rel and $\delta_1$ with different test identities.} The legends illustrate how performance shifts as the test data size increases from 2K to 10K.}
    \label{fig:abs_rel}
\end{wrapfigure}

\textbf{Summary of failure modes.} Combining Tables~\ref{tab:resillum} and \ref{tab:scenetypes}, we observe three qualitatively distinct failure modes for the ROVR-trained single-domain models:
(i)~\emph{photometric collapse} under night and rainy conditions, where low contrast causes the model to emit an unstructured, low-variance depth map;
(ii)~\emph{geometric confusion} in urban scenes with heavy clutter, where the model correctly identifies the presence of objects but predicts their distance off by factors of 1.5--2;
(iii)~\emph{range saturation} in open rural and highway scenes, where the model under-predicts the depth of distant structures because the upper tail of its training distribution is thinner than that of its deployment distribution.
Each mode calls for a different remedy (better low-light pretraining, explicit multi-scale reasoning, and broader range coverage, respectively), and identifying them is possible only because ROVR stratifies the evaluation by condition and scene rather than reporting a single aggregate number.

\textbf{Influence of test set scale.} Fig.~\ref{fig:abs_rel} shows that as the number of test identities increases from 2K to 10K, performance curves remain stable, indicating that the ROVR evaluation is unbiased and does not depend on test set size. This stability confirms the dataset's reliability as a benchmark: a small evaluation run gives a faithful picture of the full-split behaviour, which is important for practitioners who want to iterate quickly before committing to a full-scale evaluation. The stability also suggests that the test split is large enough that statistical noise is not the bottleneck for observed performance gaps---differences between methods can be attributed to modelling rather than sampling.

\textbf{Ground-truth density ablation.}
A natural question is whether the sparse LiDAR ground truth in ROVR is sufficient for training competitive depth models. To investigate this, we progressively thin the projected depth maps to 75\%, 50\%, 25\%, and 10\% of their original density and retrain VA-DepthNet under each setting. Thinning is performed by Bernoulli-masking valid pixels per frame, so that the conditional distribution over supervised pixels remains representative of the full label set.

\subsection{Discussion}
\label{sec:discussion}
The experiments above establish several design conclusions that extend beyond the ROVR benchmark itself.

\textbf{In-domain accuracy has ceased to be a useful headline number.} Every KITTI-trained model we tested scored $\delta_1 > 0.98$ on KITTI, but the same models fell below $0.25$ on ROVR. A single $\delta_1$ number therefore tells the reader almost nothing about deployment behaviour. We recommend that future papers report both an in-domain number and at least one out-of-distribution transfer number---ROVR is one natural choice because it is explicitly constructed to expose the failure modes of benchmark-saturated models.

\textbf{Scene complexity dominates sensor density.} The performance gap between highway and urban scenes (Table~\ref{tab:scenetypes}) is larger than the gap induced by thinning LiDAR density by a factor of four (density ablation). This suggests that, for practical deployment, investing in broader scene coverage---which is inexpensive with commodity hardware---dominates investing in denser LiDAR. The architectural implication is that future depth models should focus on scene-level reasoning (multi-scale objects, occlusion, clutter) rather than on the last percentage of pixel-wise regression accuracy.

\textbf{Sparse supervision is compatible with competitive accuracy.} The density ablation demonstrates that trained models retain most of their accuracy down to 25--50\% of the native label density. This validates the ROVR acquisition philosophy and, importantly, suggests that future datasets can employ even cheaper sensors if the pose and calibration pipeline remains tight. The bottleneck for further scale is therefore geographic and temporal coverage, not sensor density.

\textbf{Condition-specific failure modes call for condition-specific training protocols.} The three failure modes identified in the ablation (photometric collapse, geometric confusion, range saturation) are each best addressed by a different intervention. A practical recommendation from this work is that training recipes for driving depth should incorporate condition-aware balancing---for example, up-weighting night and rainy samples during the final fine-tuning phase---rather than using uniform sampling. We leave the design of such recipes as future work, but we release the condition tags alongside the dataset so that researchers can build on this immediately.

\begin{figure}[!t]
\centering
\setlength{\tabcolsep}{0.8pt}
\renewcommand{\arraystretch}{1.0}
\newcommand{\qualimg}[1]{\includegraphics[width=\linewidth]{#1}}
\newcommand{\qualimgcap}[2]{%
    \begin{minipage}{\linewidth}
        \centering
        \includegraphics[width=\linewidth]{#1}\\[-1pt]
        {\scriptsize #2}
    \end{minipage}%
}
\newcommand{\qualrow}[6]{%
    {\scriptsize\makecell[l]{#1}} &
    \qualimg{#2} &
    \qualimg{#3} &
    \qualimg{#4} &
    \qualimg{#5} &
    \qualimg{#6} \\
}
\newcommand{\qualrowcap}[6]{%
    {\scriptsize\makecell[l]{#1}} &
    \qualimgcap{#2}{Image} &
    \qualimgcap{#3}{GT} &
    \qualimgcap{#4}{VA-Depthnet} &
    \qualimgcap{#5}{IEBins} &
    \qualimgcap{#6}{DCDepth} \\
}
\begin{adjustbox}{width=\linewidth,max totalheight=0.72\textheight,keepaspectratio}
\begin{tabular}{@{}>{\raggedright\arraybackslash}m{0.075\textwidth}*{5}{m{0.183\textwidth}}@{}}
    \qualrow{Highway\\(Night)}
        {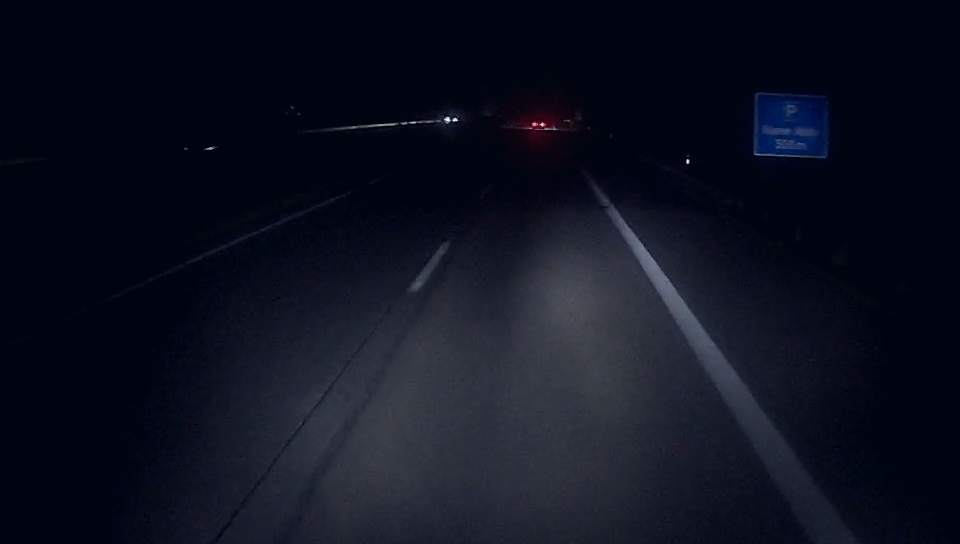}
        {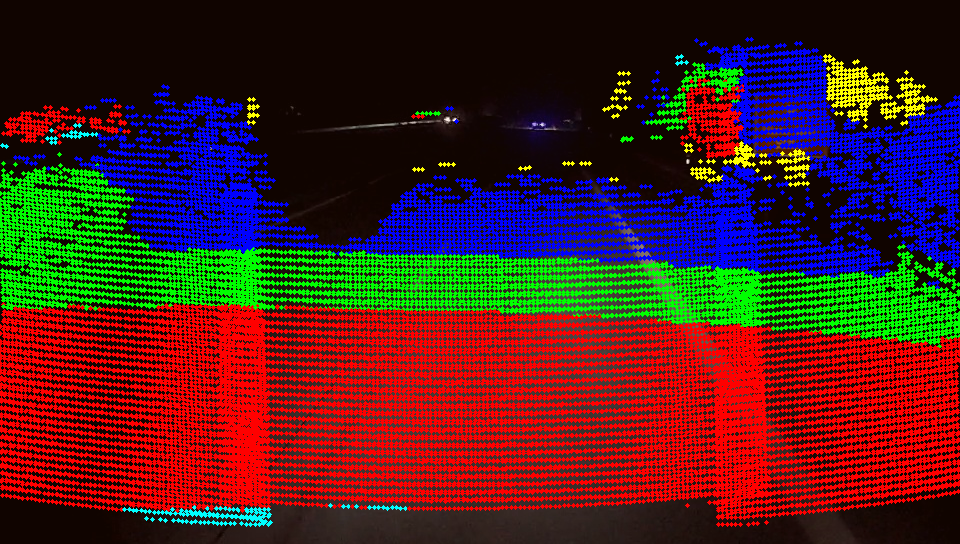}
        {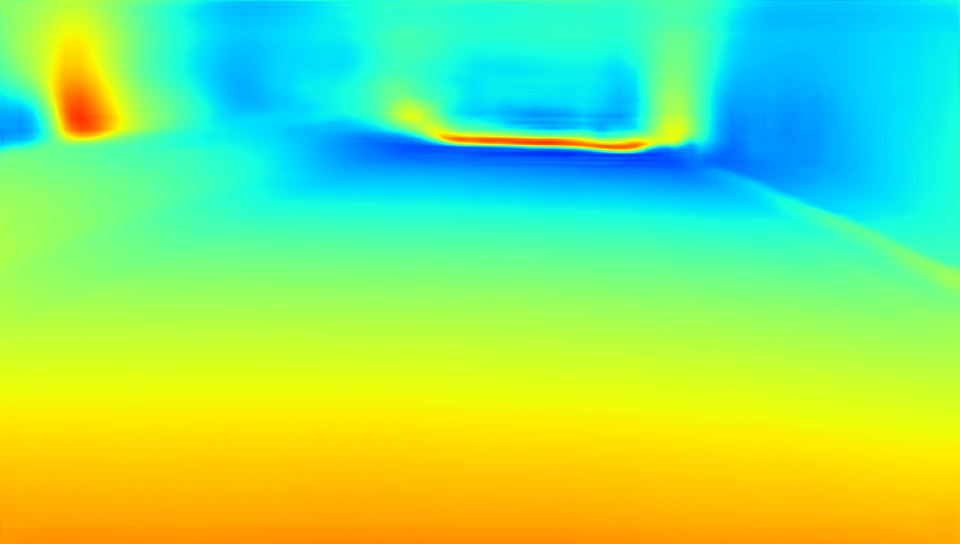}
        {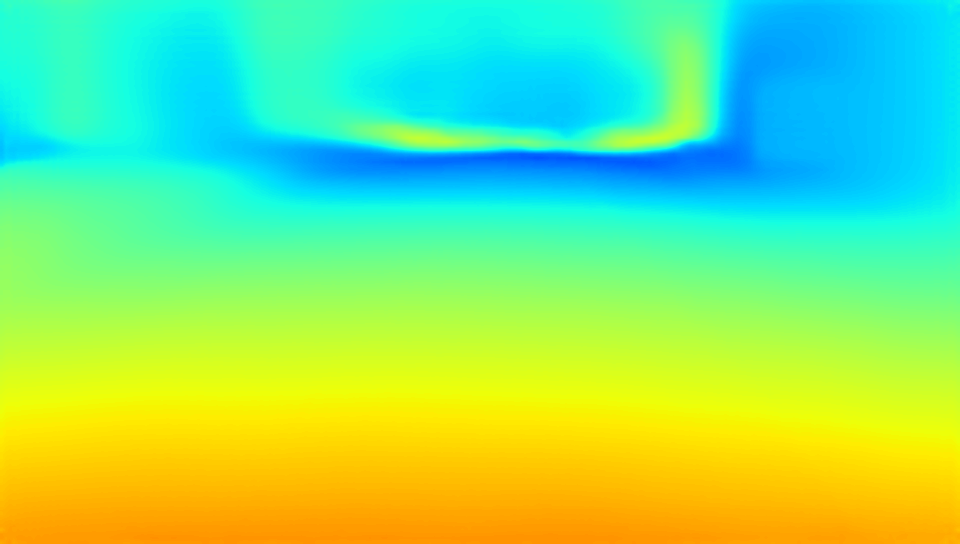}
        {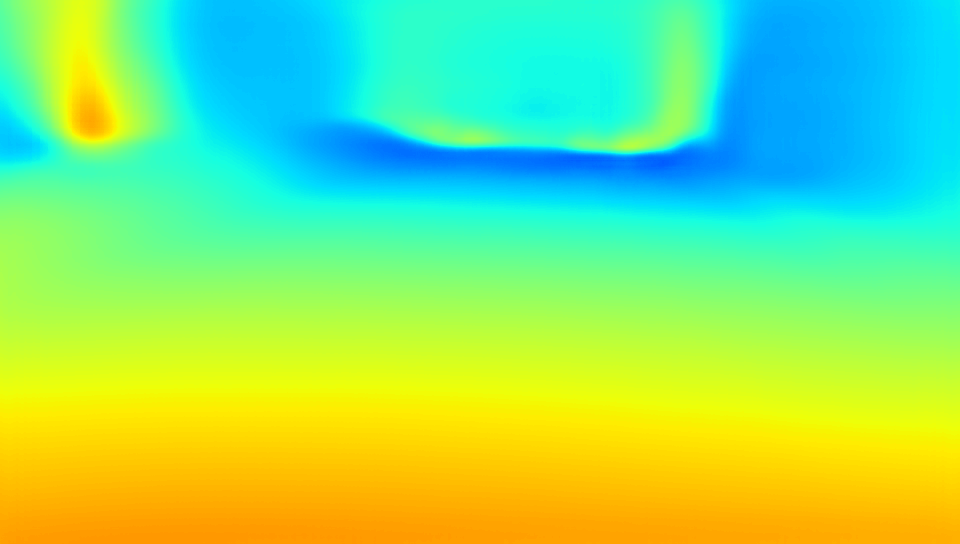}
    \qualrow{Highway\\(Normal)}
        {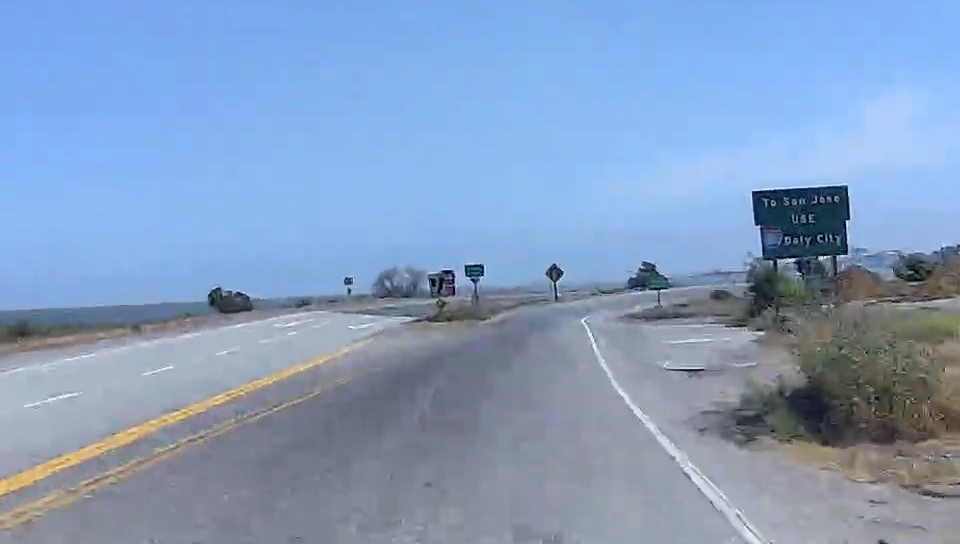}
        {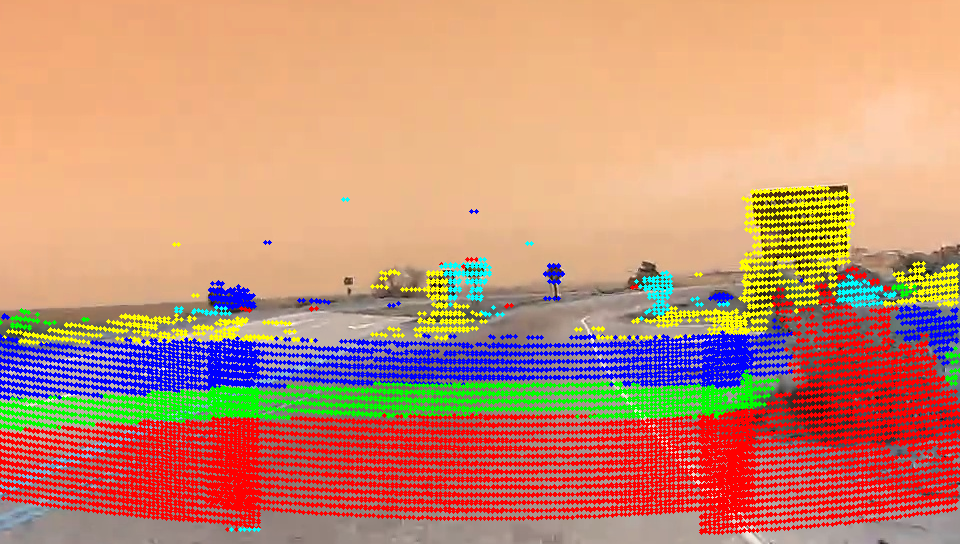}
        {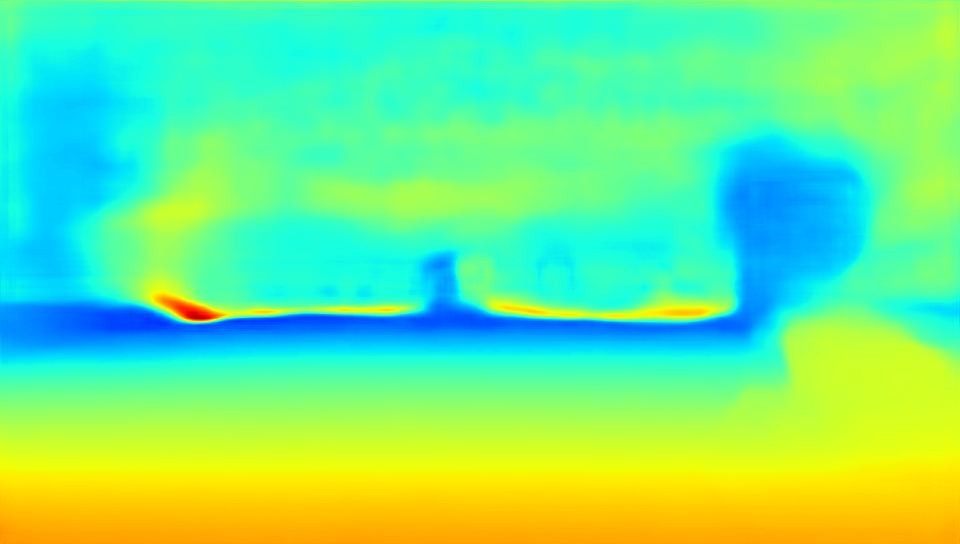}
        {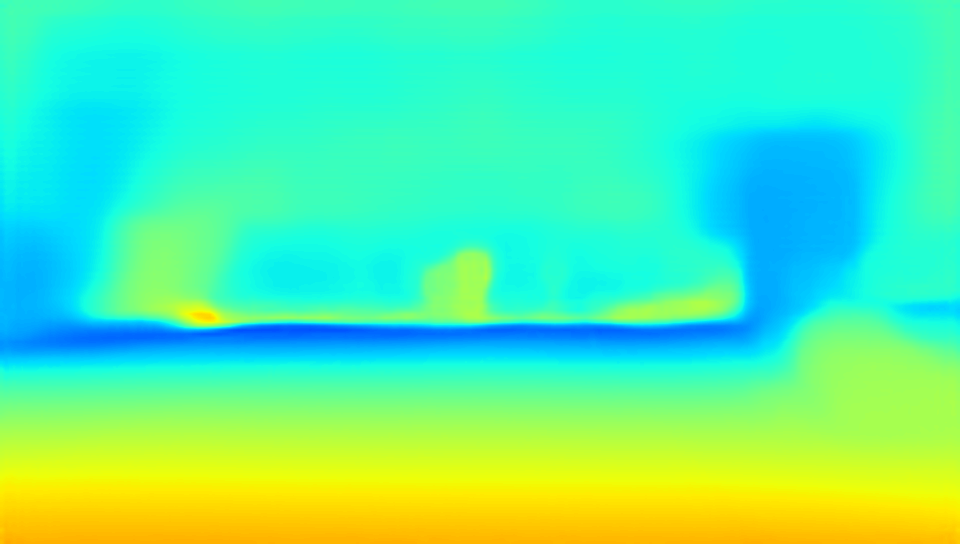}
        {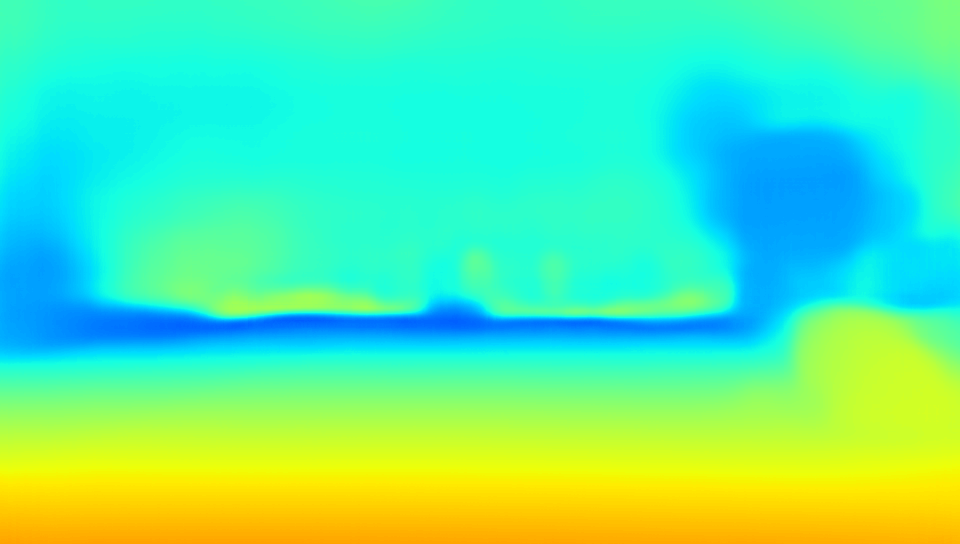}
    \qualrow{Highway\\(Rainy)}
        {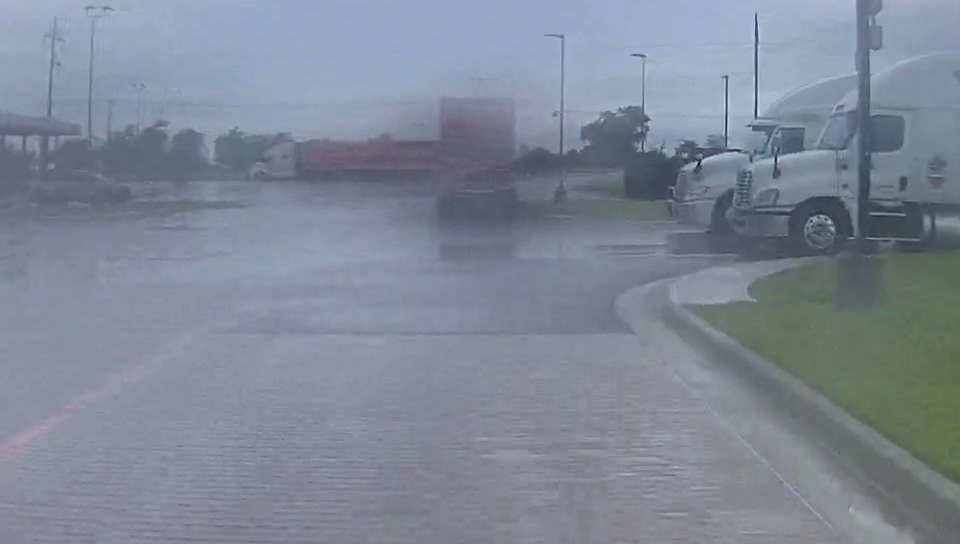}
        {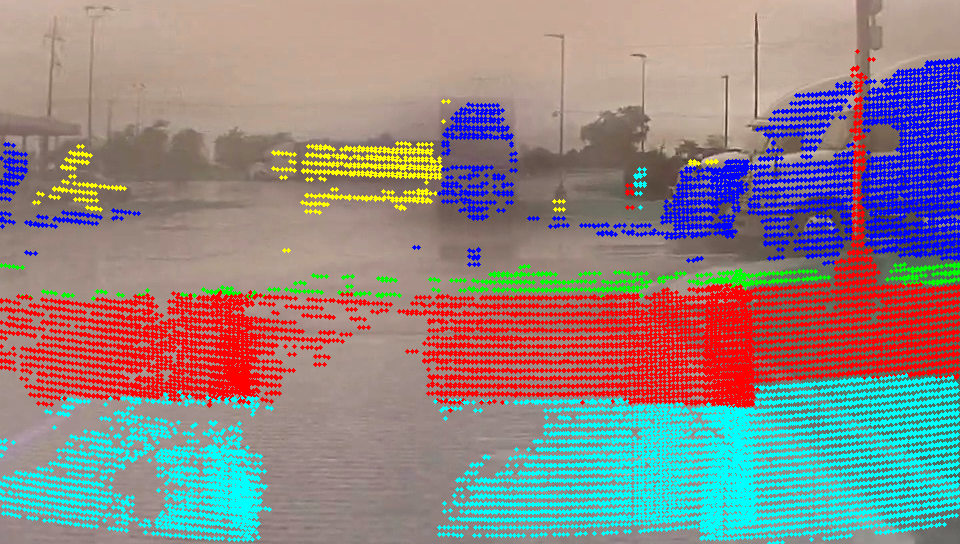}
        {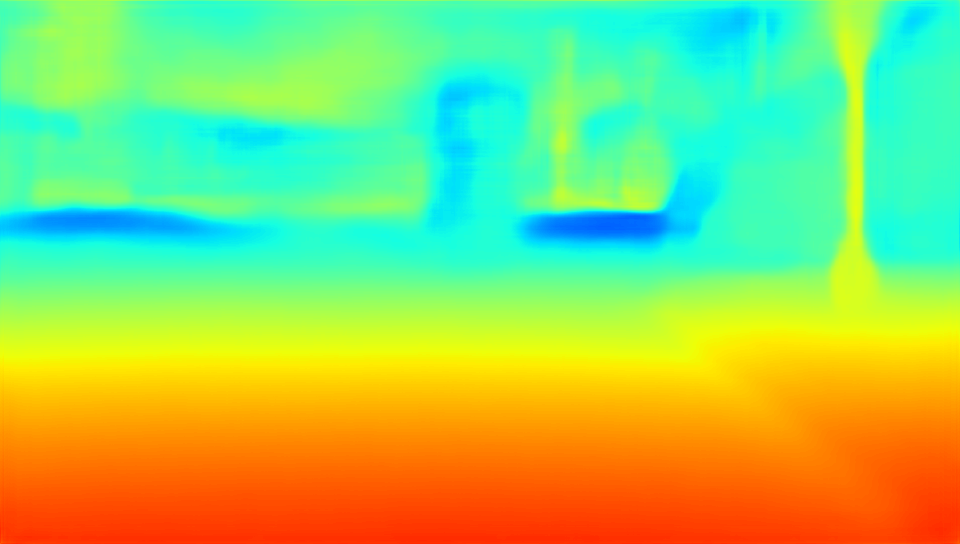}
        {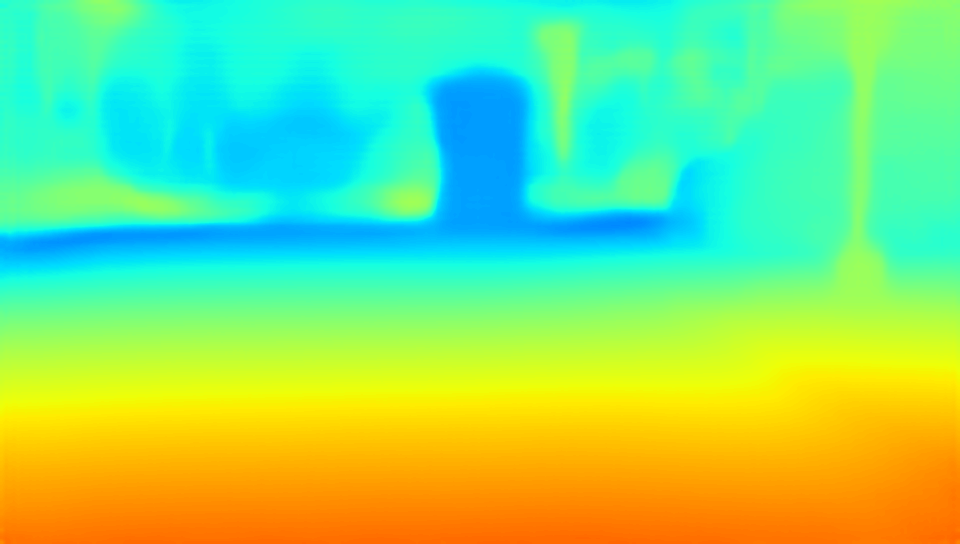}
        {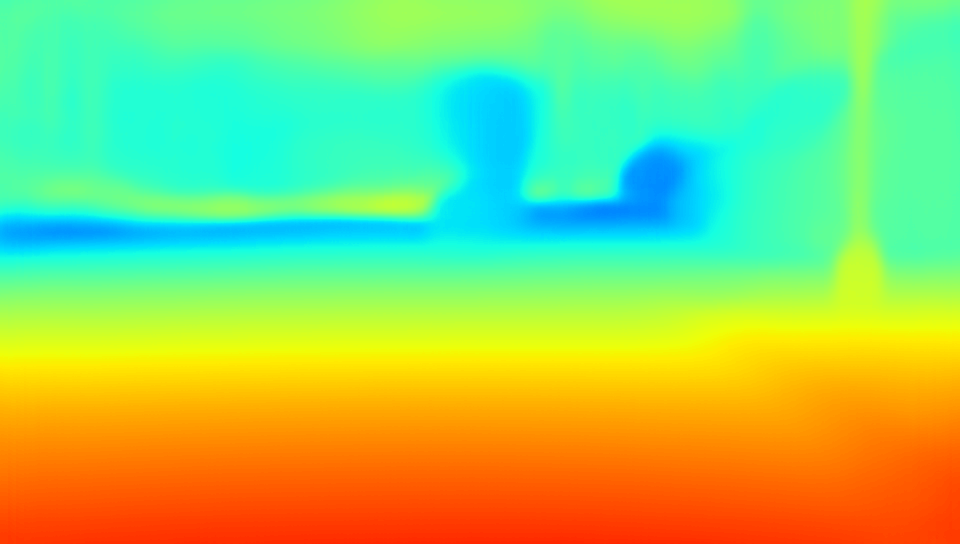}
    \qualrow{Rural\\(Normal)}
        {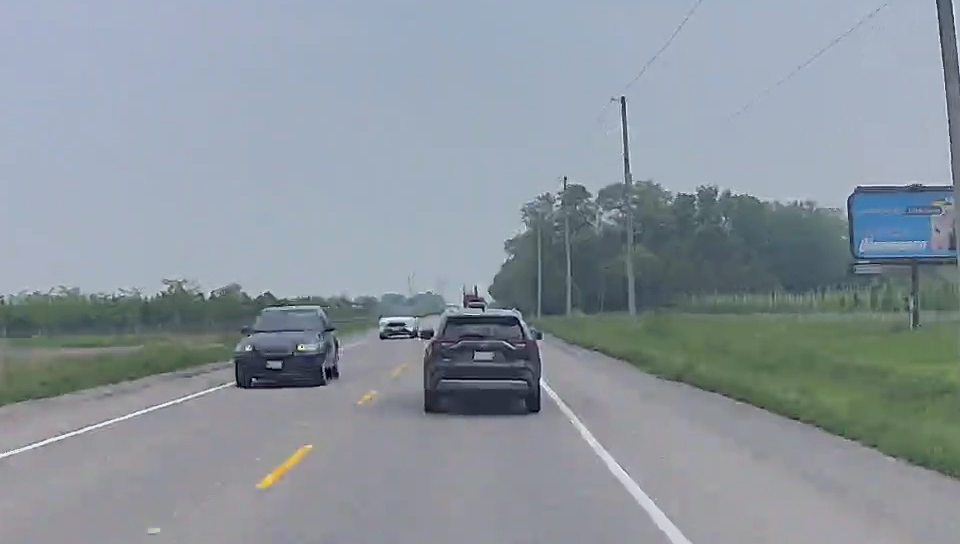}
        {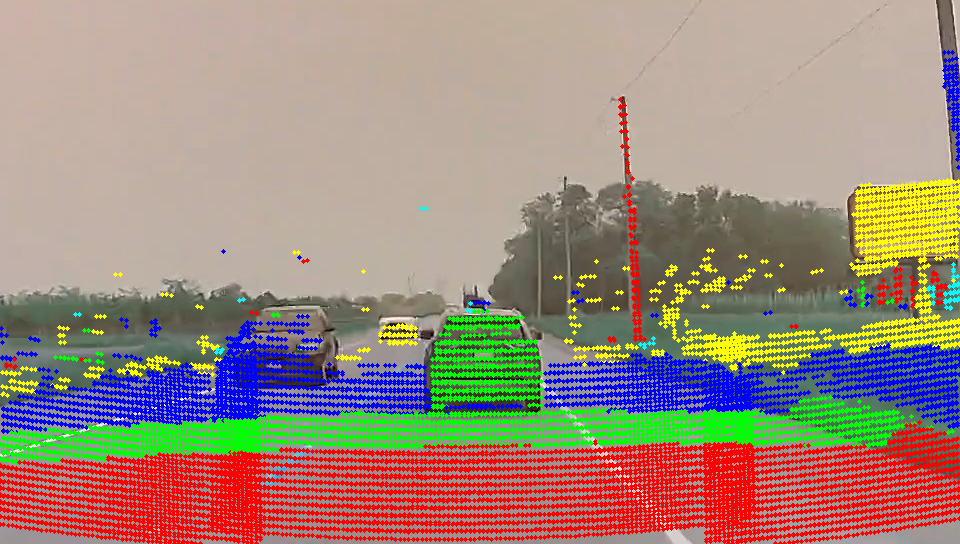}
        {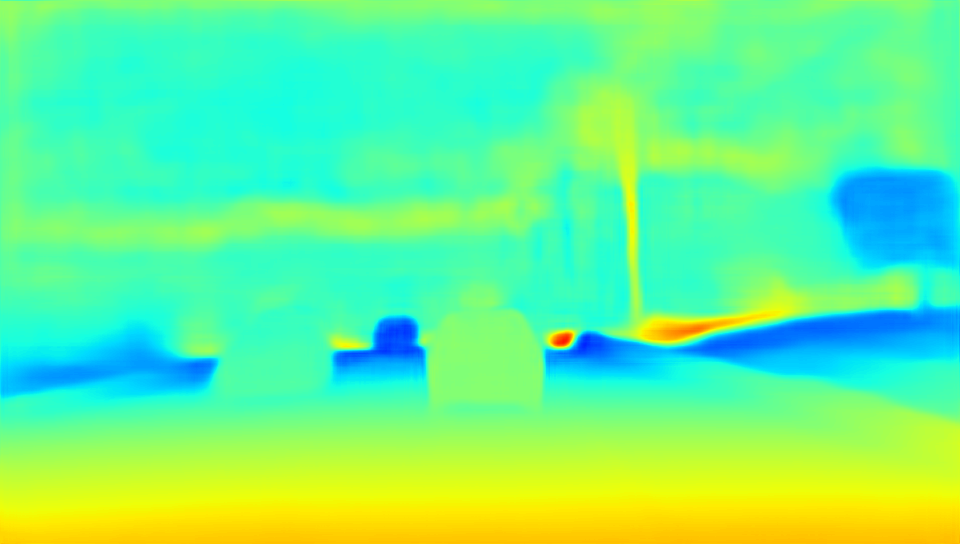}
        {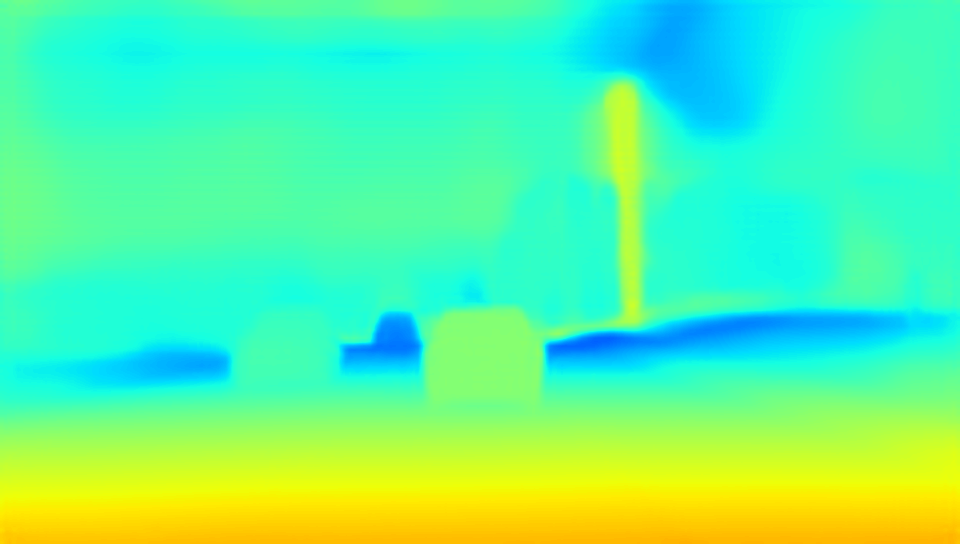}
        {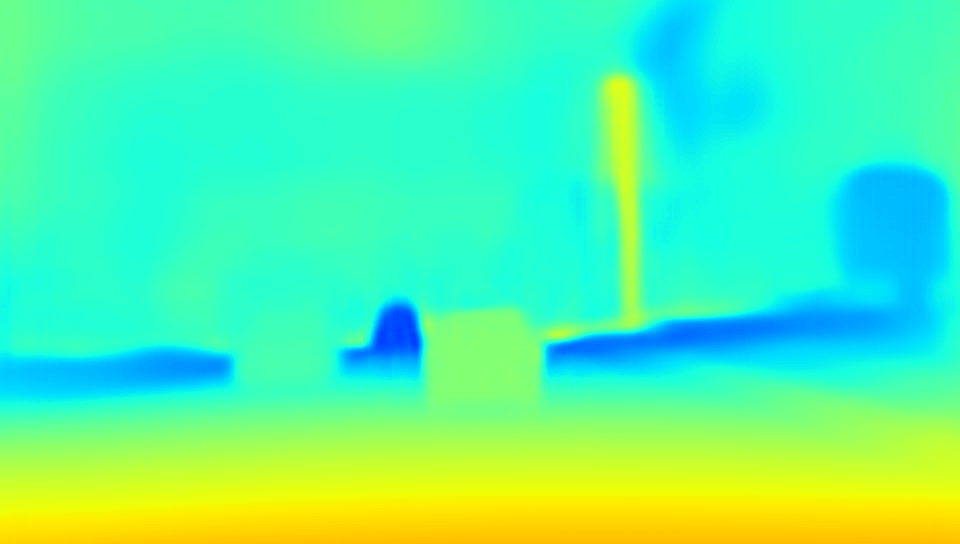}
    \qualrow{Rural\\(Rainy)}
        {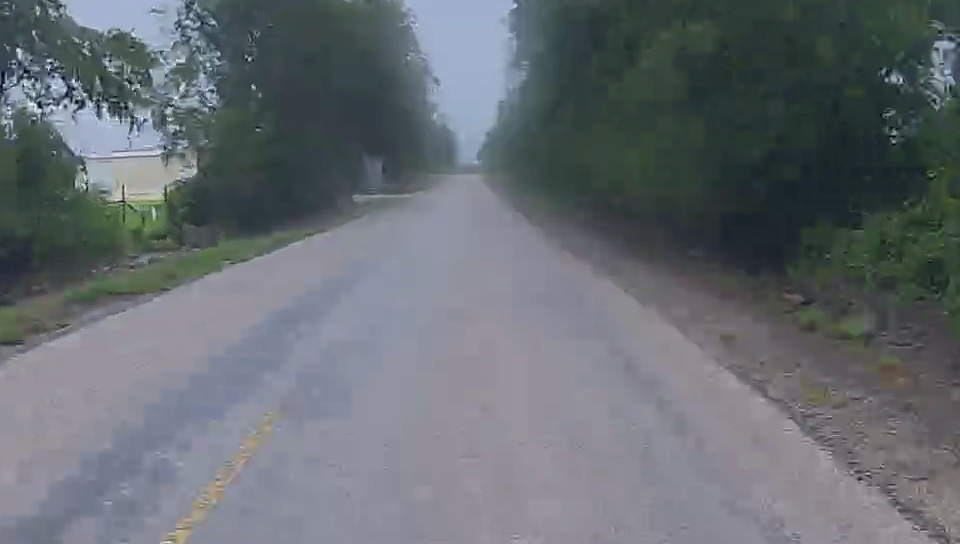}
        {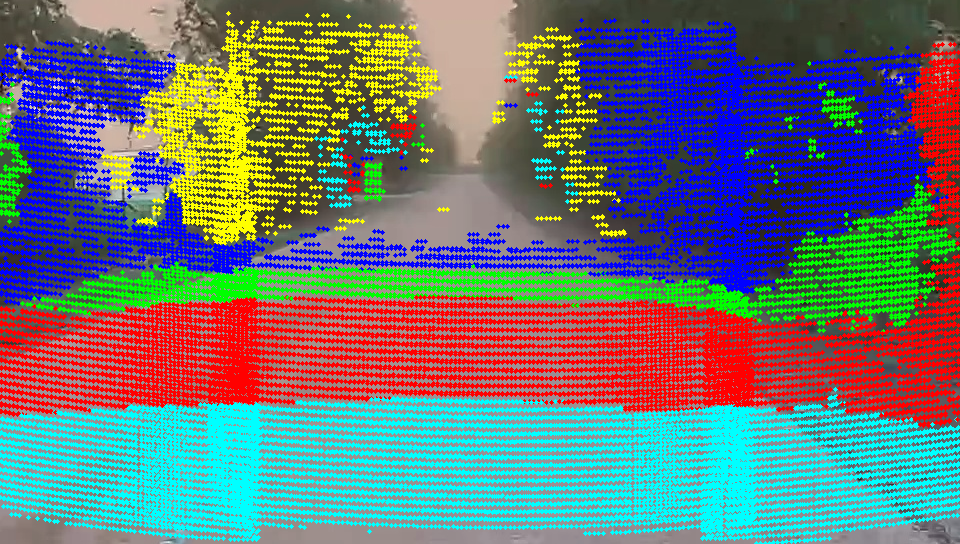}
        {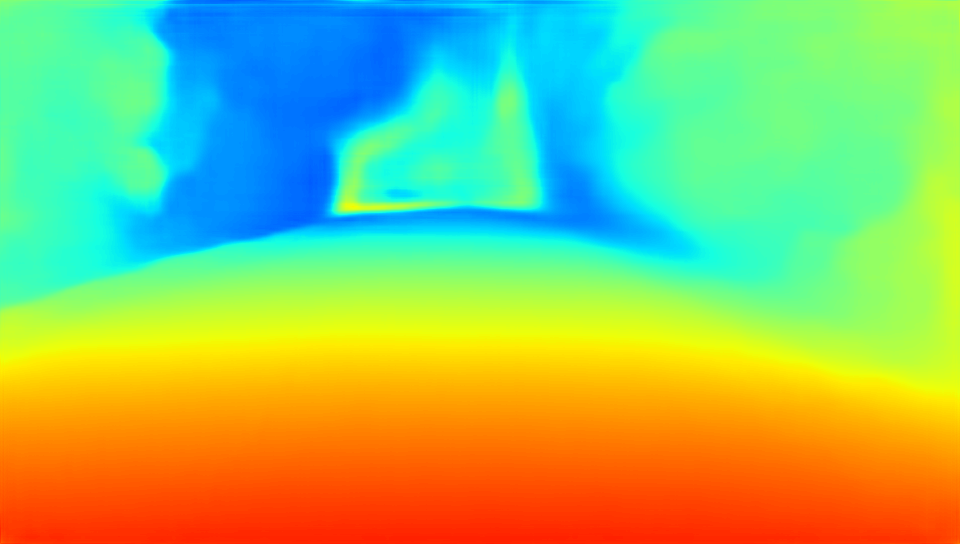}
        {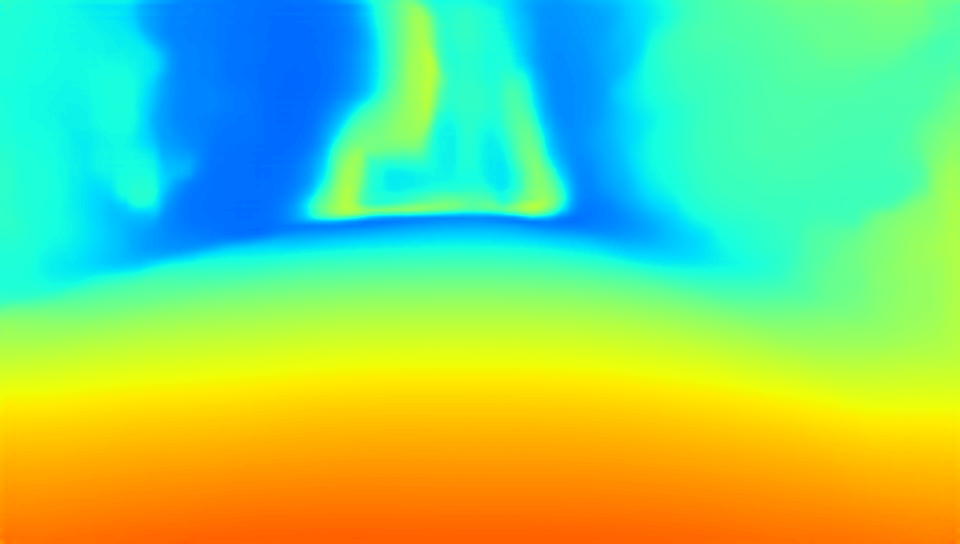}
        {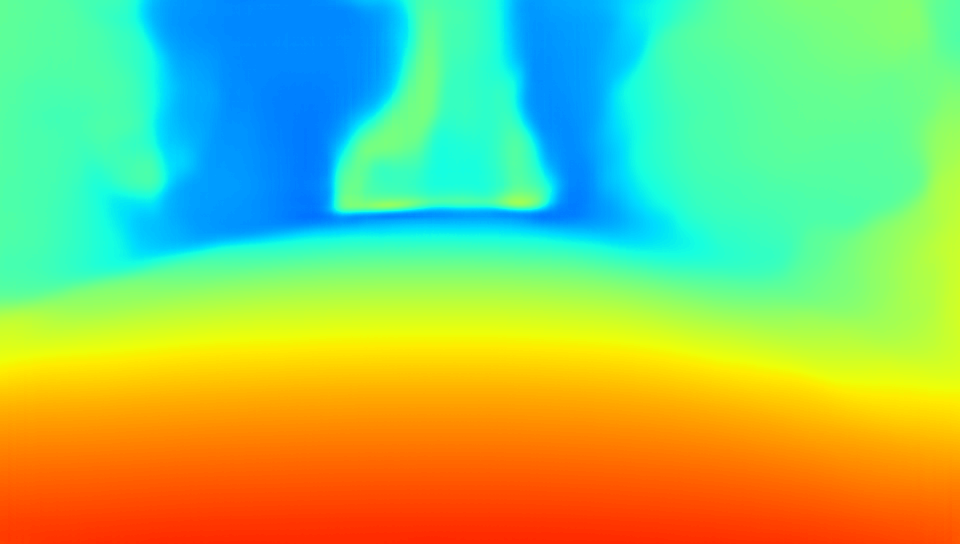}
    \qualrow{Urban\\(Night)}
        {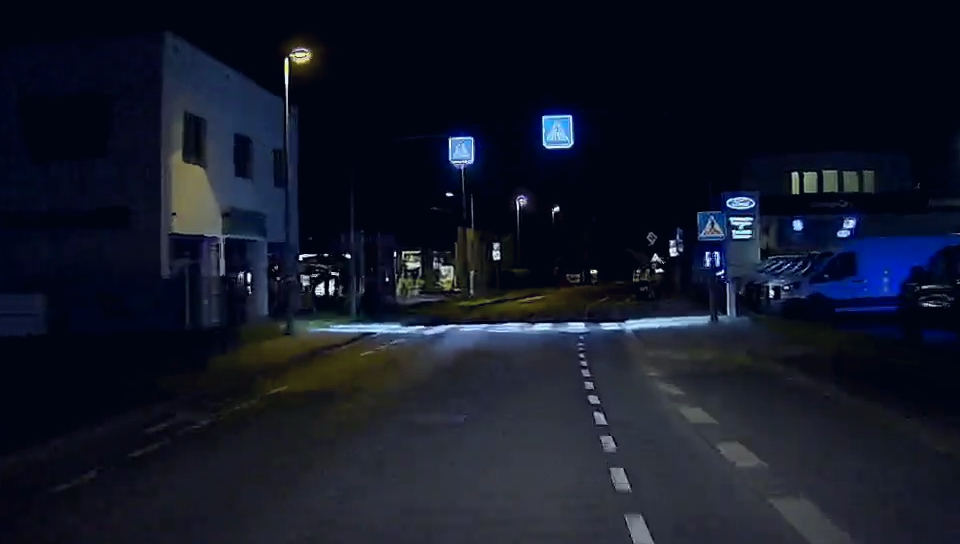}
        {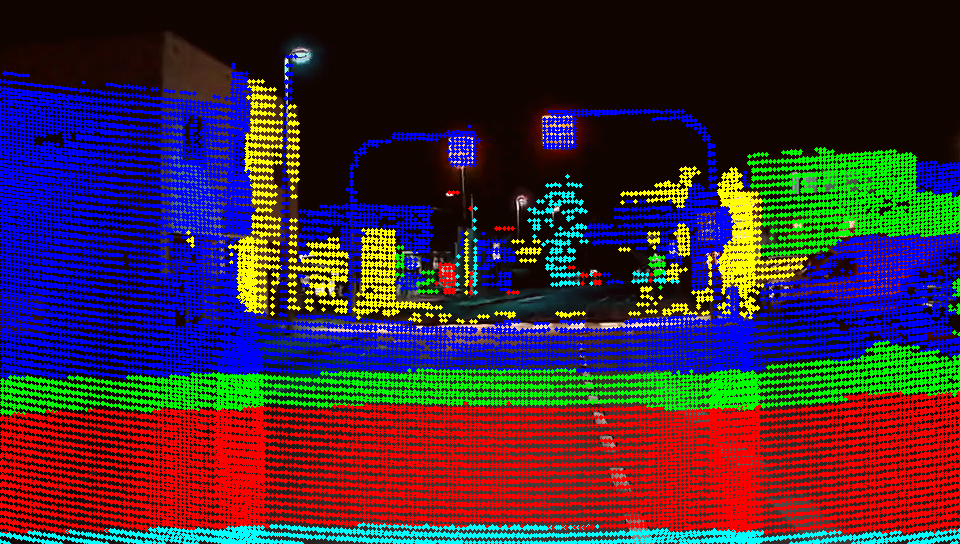}
        {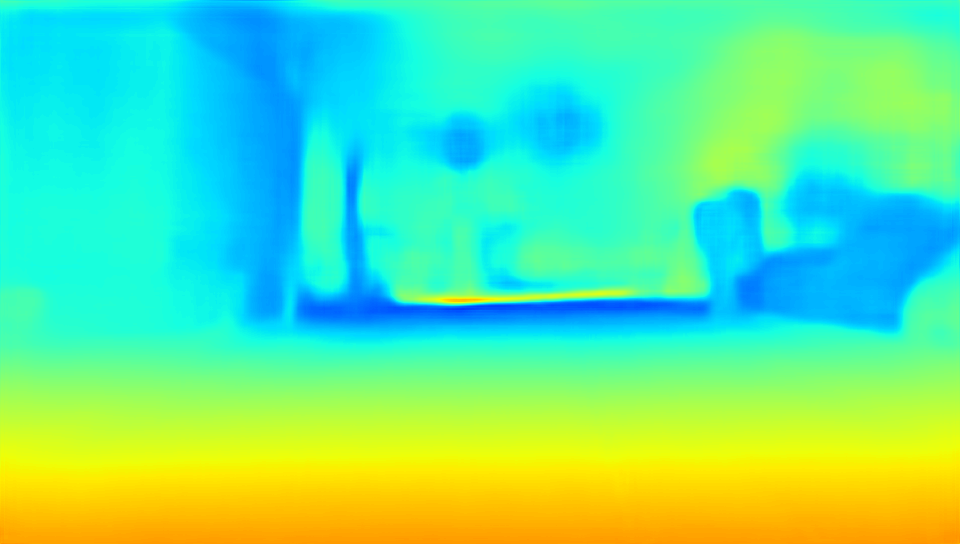}
        {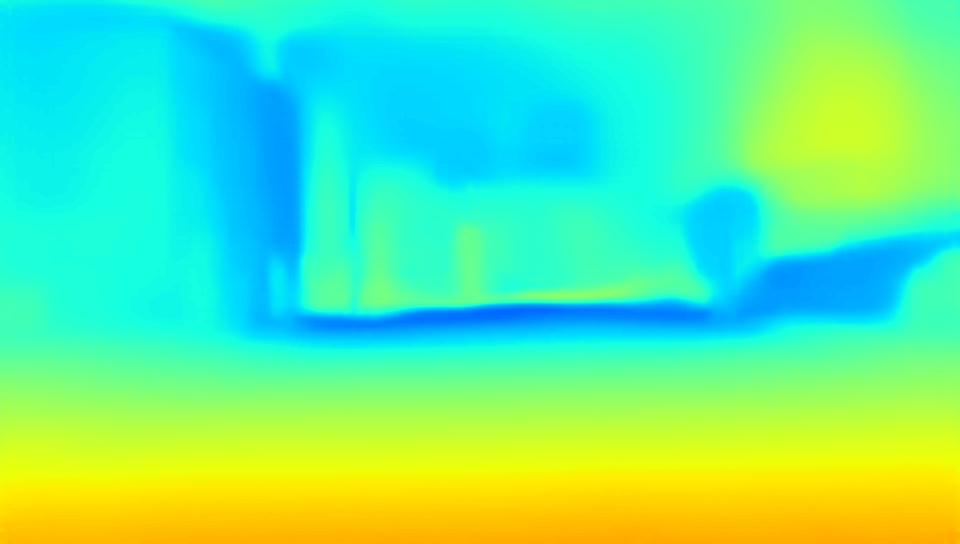}
        {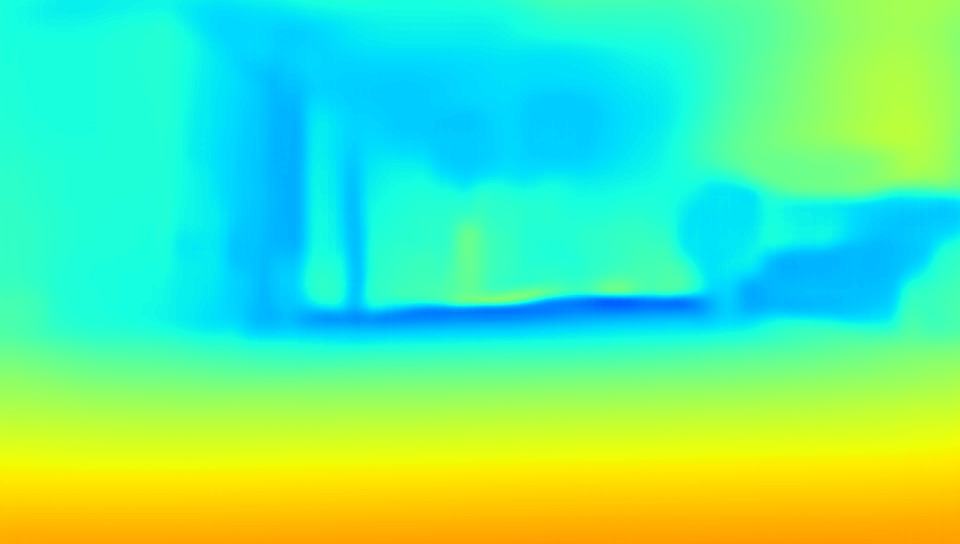}
    \qualrow{Urban\\(Normal)}
        {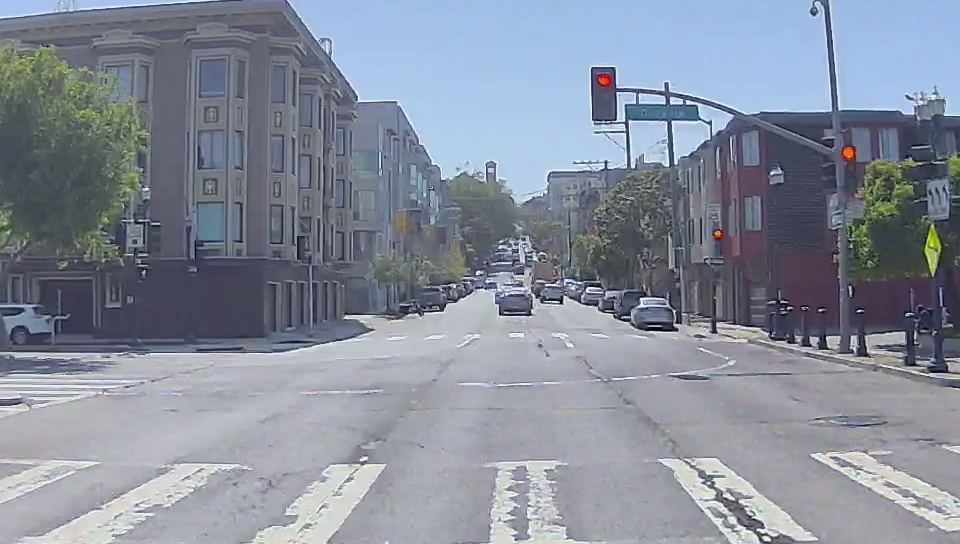}
        {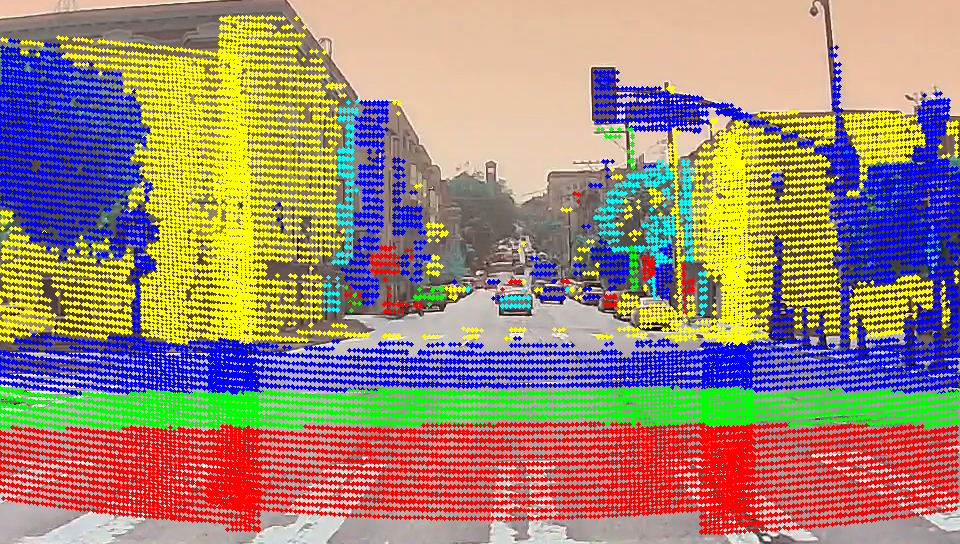}
        {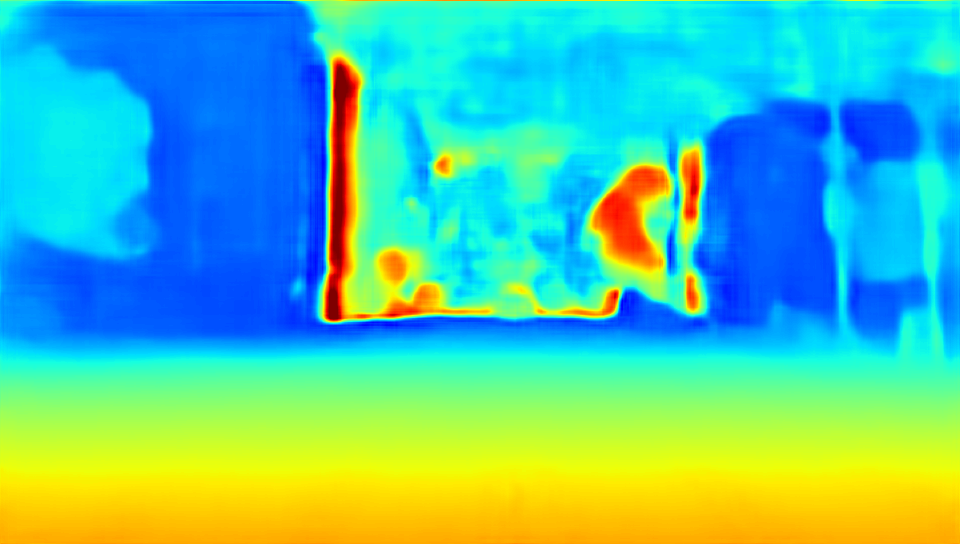}
        {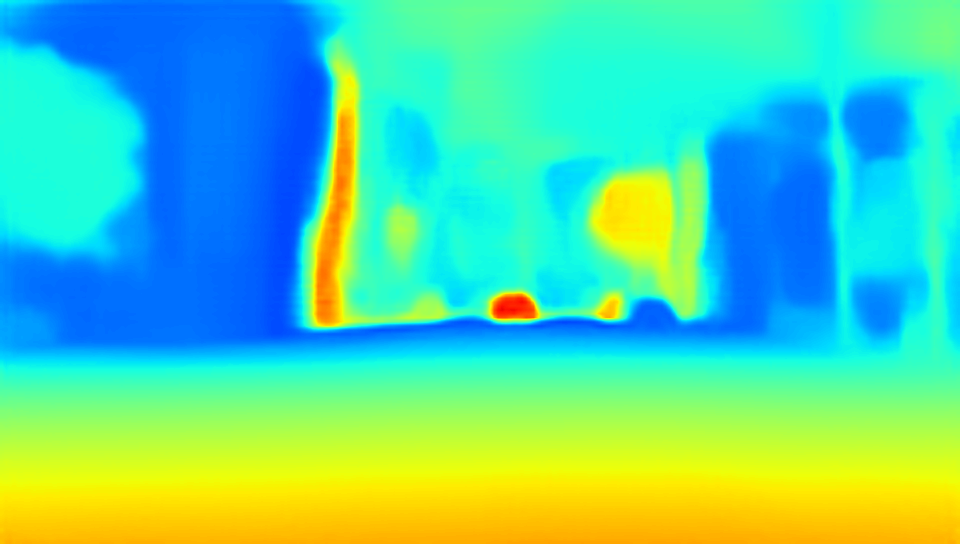}
        {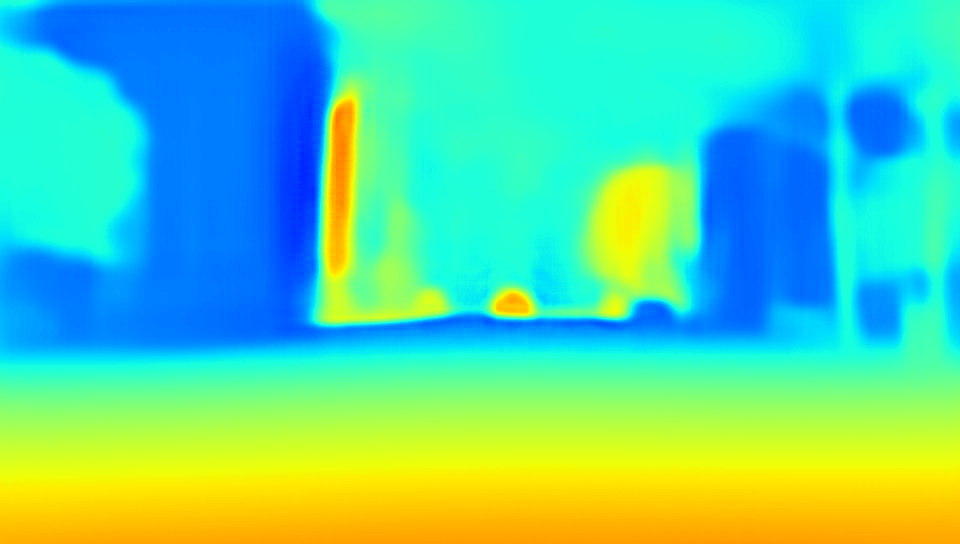}
    \qualrowcap{Urban\\(Rainy)}
        {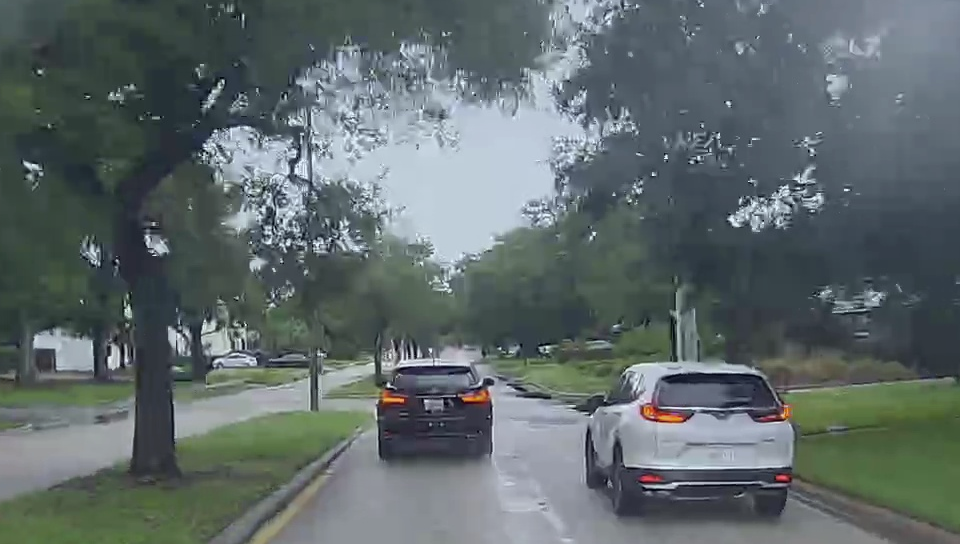}
        {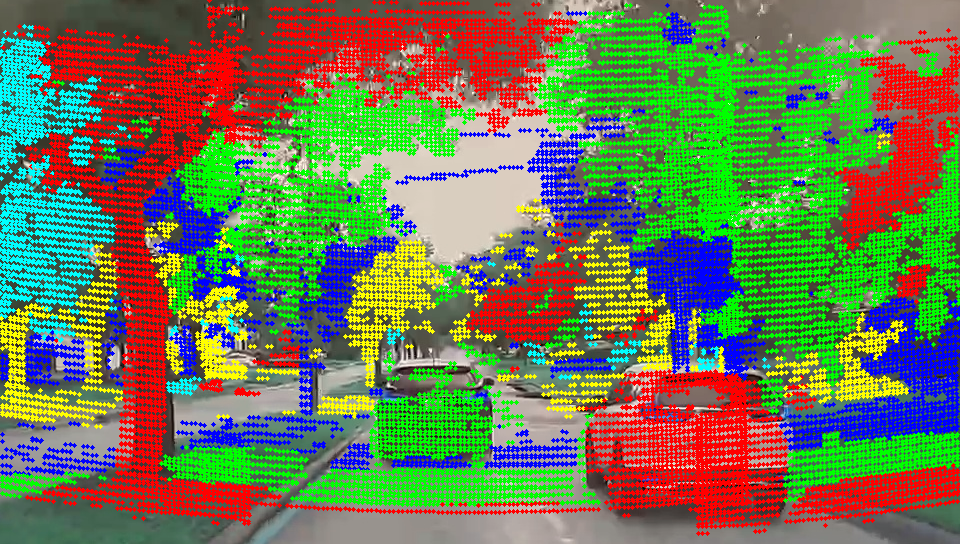}
        {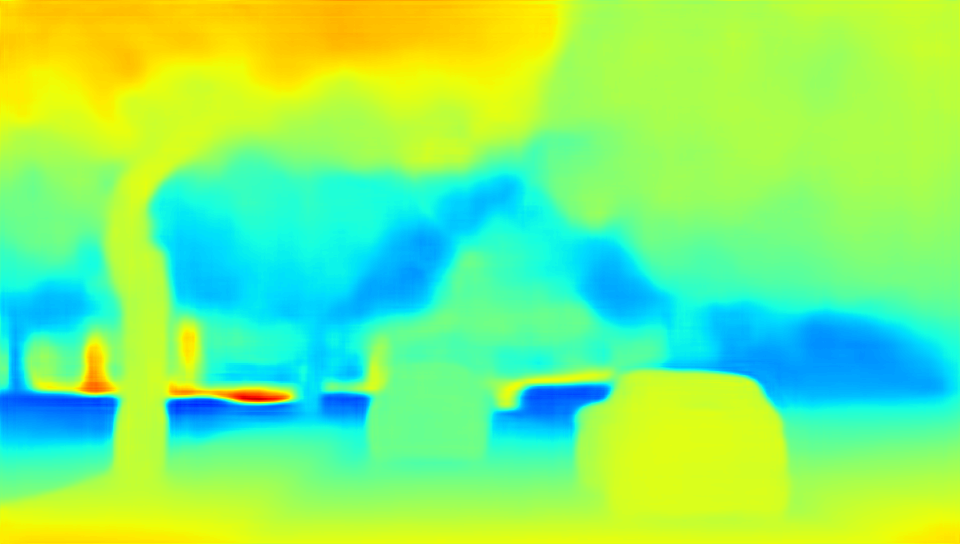}
        {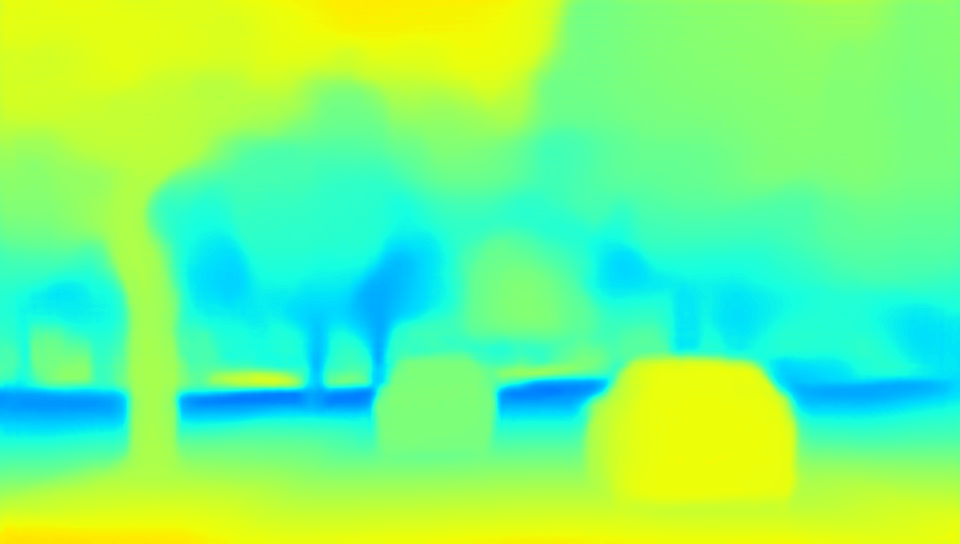}
        {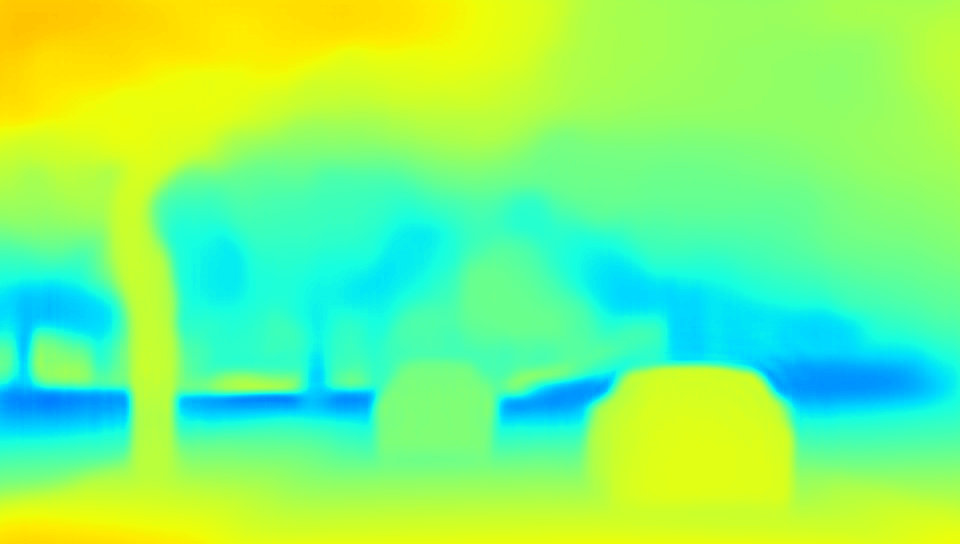}
\end{tabular}
\end{adjustbox}
\vspace{-1mm}
\caption{\textbf{Qualitative comparisons} of depth estimation results across different scenarios (highway, rural, urban) and weather conditions (night, normal, rainy).}
\label{fig:qualitative}
\vspace{-2mm}
\end{figure}

\subsection{Qualitative Results}

Fig.~\ref{fig:qualitative} presents qualitative depth predictions across scene types (highway, rural, urban) and conditions (normal, night, rainy), corresponding to the settings in Tables~\ref{tab:resillum} and~\ref{tab:scenetypes}.
Adverse weather and nighttime conditions visibly degrade prediction quality and reduce ground-truth density, while urban environments introduce clutter, occlusions, and multi-scale structures that cause clear failure modes. These examples illustrate that ROVR captures real-world driving challenges that expose the limitations of existing depth estimation methods. The qualitative gallery is organized so that each row shares an RGB input and shows the predictions of the three single-domain baselines side by side; this layout makes the three failure modes of Sec.~\ref{sec:experiments} directly legible, since photometric collapse appears as loss of structure, geometric confusion as scale inconsistency between nearby objects, and range saturation as flattened depth in the far field.

\section{Conclusion and Future Work}
\label{sec:conclusion}

We introduced ROVR, a large-scale, diverse, and cost-efficient depth dataset for real-world driving. Covering varied scenes, weather, and illumination conditions across three continents, ROVR provides sparse yet statistically sufficient LiDAR ground truth for robust training and challenging evaluation. Our experiments show that state-of-the-art depth models degrade substantially on ROVR, and even models trained directly on it remain far from saturation. 

By releasing both the benchmark and the documented sensor suite, ROVR aims to promote depth estimation methods that are more robust to complex real-world environments while lowering the barrier for reproducible data collection.

\paragraph{Limitations.}
The current release focuses on forward-facing monocular depth with sparse LiDAR ground truth. Solid-state LiDAR produces fewer returns on reflective, transparent, and low-reflectance surfaces, creating systematic gaps that may limit analysis of such failure cases. Although ROVR spans three continents, polar, high-altitude, and severe-weather conditions remain under-represented, and the rain subset mainly covers moderate precipitation. The single-camera forward-facing layout preserves the platform's low cost, but limits direct applicability to 360$^{\circ}$ perception and multi-camera fusion; thus, ROVR complements rather than replaces surround-view datasets such as nuScenes and Waymo.

\section*{Declaration of competing interest}

The authors declare that they have no known competing financial interests or personal relationships that could have appeared to influence the work reported in this paper.

\section*{Declaration of Use of Generative AI}
The authors declare that generative AI tools were used only for language polishing and grammar checking. All scientific content, including the methodology, experiments, and conclusions, was developed and verified by the authors.

\section*{Credit author statement}

Xianda Guo: Investigation, Writing -- Original Draft; Writing -- Review \& Editing, Conceptualization. 
Ruijun Zhang: Software, Data Curation, Validation. 
Yiqun Duan: Writing -- Review \& Editing.
Ruilin Wang: Software, Visualization. 
Matteo Poggi: Writing -- Review \& Editing.
Keyuan Zhou: Methodology, Software. 
Wenzhao Zheng: Investigation, Software, Visualization. 
Wenke Huang: Writing -- Review \& Editing.
Gangwei Xu: Writing -- Review \& Editing.
Yanlun Peng: Methodology, Software. 
Yuan Si: Data Curation, Validation. 
Qin Zou: Writing -- Review \& Editing, Supervision.

\section*{Acknowledgements}
This work was supported by ROVR Labs, Inc. We thank the vehicle operators and the ROVR data-engineering team for the fleet coordination, calibration, and privacy audit, and the anonymous reviewers for their constructive feedback.

\bibliography{main}

\end{document}